\documentclass[10pt,twocolumn,letterpaper]{article}

\usepackage{wacv}
\usepackage{times}
\usepackage{epsfig}
\usepackage{graphicx}
\usepackage{amsmath}
\usepackage{amssymb}
\usepackage{booktabs}
\usepackage{subcaption,siunitx,booktabs}
\usepackage{color}
\usepackage{sidecap}
\definecolor{atomictangerine}{rgb}{0.8, 0.2, 0.1}
\definecolor{turq}{rgb}{0.0, 0.5, 0.5}
\definecolor{darkturq}{rgb}{0.0, 0.4, 0.4}
\definecolor{bright}{rgb}{0.8, 0.1, 0}
\definecolor{darkgray}{gray}{0.3}
\definecolor{mahogany}{rgb}{0.6, 0.05, 0.05}
\definecolor{myblue}{rgb}{0.3,0.05,0.9}

\definecolor{olive}{rgb}{0.537, 0.627, 0.318}
\definecolor{green}{rgb}{0.22, 0.463, 0.114}
\definecolor{grey}{rgb}{0.4, 0.4, 0.4}
\definecolor{blue}{rgb}{0.435, 0.659, 0.863}
\definecolor{pink}{rgb}{0.761, 0.482, 0.627}
\definecolor{darkpink}{rgb}{0.561, 0.282, 0.427}

\newcommand\ignore[1]{}

\newcommand{\tildeapprox}{{\raise.17ex\hbox{$\scriptstyle\sim$}}}
\newcommand{\secref}[1]{Section~\ref{#1}}
\newcommand{\figref}[1]{Figure~\ref{#1}}

\renewcommand{\eqref}[1]{Eq.~(\ref{#1})}

\renewcommand\vec[1]{\mathbf{#1}}
\renewcommand{\a}{\vec{a}}
\newcommand{\x}{\vec{x}}
\newcommand{\y}{y}
\newcommand{\w}{\vec{w}}
\newcommand{\Y}{\mathcal{Y}}
\newcommand{\VE}{V}
\newcommand{\AEx}{S}
\newcommand{\FV}{\w_{\VE}}
\newcommand{\FA}{\w_{\AEx}}
\newcommand{\h}{\vec{h}}

\newcommand{\DRAGON}{\textsc{Dragon}}
\newcommand{\SmoothTail}{Smooth-Tail}
\newcommand{\TwoLevel}{Two-Level}

\def\wacvPaperID{397} %

\wacvfinalcopy %

\ifwacvfinal
\def\assignedStartPage{1} %
\fi

\ifwacvfinal
\usepackage[pagebackref=true,breaklinks=true,colorlinks,bookmarks=false]{hyperref}
\else
\usepackage[pagebackref=true,breaklinks=true,colorlinks,bookmarks=false]{hyperref}
\fi

\ifwacvfinal
\else
\fi

\begin{document}

\title{From Generalized zero-shot learning to long-tail with class descriptors} 

\author{\textbf{Dvir Samuel}\textsuperscript{1}
\textbf{Yuval Atzmon}\textsuperscript{2}
\textbf{Gal Chechik}\textsuperscript{1,2} \\
\textsuperscript{1}Bar-Ilan University, Ramat Gan, Israel \\
\textsuperscript{2}NVIDIA Research, Tel Aviv, Israel \\
dvirsamuel@gmail.com, yatzmon@nvidia.com, gal.chechik@biu.ac.il}

\maketitle

\begin{abstract}
Real-world data is predominantly unbalanced and long-tailed, but deep models struggle to recognize rare classes in the presence of frequent classes. Often, classes can be accompanied by side information like textual descriptions, but it is not fully clear how to use them for learning with unbalanced long-tail data. Such descriptions have been mostly used in (Generalized) Zero-shot learning (ZSL), suggesting that ZSL with class descriptions may also be useful for long-tail distributions. 

We describe \DRAGON{}, a late-fusion architecture for long-tail learning with class descriptors.
It learns to (1)  correct the bias towards head classes on a sample-by-sample basis; and (2) fuse information from class-descriptions to improve the tail-class accuracy. We also introduce new benchmarks CUB-LT, SUN-LT, AWA-LT for long-tail learning with class-descriptions, building on existing learning-with-attributes datasets and a version of Imagenet-LT with class descriptors.
\DRAGON{} outperforms state-of-the-art models on the new benchmark. It is also a new SoTA on existing benchmarks for GFSL with class descriptors (GFSL-d) and standard (vision-only) long-tailed learning ImageNet-LT, CIFAR-10, 100, and Places365-LT.
\end{abstract}

\section{Introduction}
\label{sec:intro}
Real-world data is predominantly unbalanced, typically following a long-tail distribution. 
From text data (Zipf's law), through acoustic noise (the $1$-over-$f$ rule) to the long-tail distribution of classes in object recognition~\cite{van2017devil}, few classes are frequently observed, while the many remaining ones are rarely encountered. 

\begin{figure}
    (a) \hspace{120pt}
    (b) \hspace{90pt}
    \centering
   \includegraphics[width=0.49\linewidth]{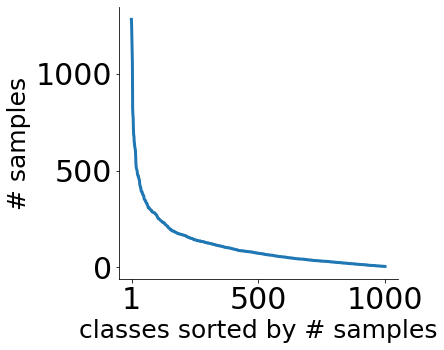}
   \includegraphics[width=0.49\linewidth]{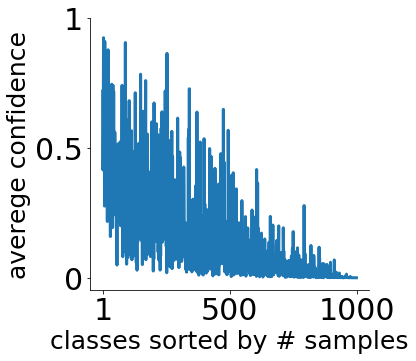}
   (c)\\
   \includegraphics[width=0.49\linewidth]{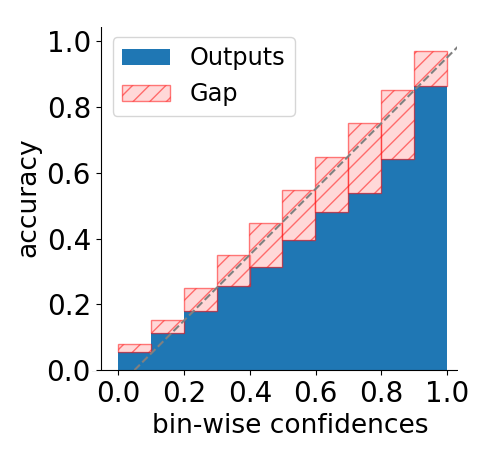}
    \caption{{Training with unbalanced data leads to a ``familiarity bias", where models are more confident and more over-confident about frequent classes~\cite{cpe,Wallace2013ImprovingCP,buda2017systematic}.
    \textbf{(a)} Class distribution of a long-tailed ImageNet~\cite{Deng2009ImageNetAL}. Classes are ordered from left to right by decreasing number of samples. \textbf{(b)} When training a ResNet-10 on ImageNet-LT, validation (and test) predictions tend to have low confidence for tail classes. We show the mean output of softmax for each class, conditioned on samples from that class. \textbf{(c)} A reliability graph for the model in b.
    Predictions are grouped based on confidence. 
    The model has larger confidence gaps (pink boxes) for more confident predictions, which usually come from head classes. 
    This result suggests that overconfidence is strongly affected by class frequency, and we can learn to correct it if the number of samples is known.}
    }
    \label{fig:unbalance}
\end{figure}

Long-tail data poses two major challenges to learning: \textit{data paucity} and \textit{data imbalance}. First, at the tail of the distribution, classes are poorly sampled and one has to use few-shot and zero-shot learning techniques. Second, when training a single model for both richly-sampled classes and poorly-sampled classes, the common classes dominate training, and as we show below, this skews prediction confidence towards rich-sampled classes. 

To address \textit{data paucity} of tail classes, note that visual examples can very often be augmented with class descriptors. Namely,  semantic information about classes given as text or attributes~\cite{DAP,Reed,LAGO,xianCVPR}.
This approach, \textit{learning with class-descriptors} has been studied mostly for zero-shot and generalized zero-shot learning~\cite{Schnfeld2019GeneralizedZL,PambalaGenerativeMW,REVISE,xianCVPR,Xian2019FVAEGAND2AF}. Here, we propose to adapt it to generalized few-shot learning (GFSL) by fusing information from two modalities. A visual classifier, expected to classify correctly head classes, and a semantic classifier, trained with per-class descriptors and is expected to classify correctly tail classes. We explain the subtleties of GZSL and GFSL in \secref{sec:related} (Related work).

To address \textit{data imbalance}, we first note that learning with unbalanced data leads to a \textbf{familiarity effect}, where models become biased to favor the more familiar, rich-sampled classes ~\cite{cpe,buda2017systematic}. Since deep models tend to be overly confident about high-confidence sample predictions \cite{Guo2017OnCO,Kull2019BeyondTS}, they become over-confident about head classes  
\cite{cpe,Wallace2013ImprovingCP,buda2017systematic}. Figure \ref{fig:unbalance} illustrates this phenomenon. It shows that a model trained on unbalanced data (\figref{fig:unbalance}a), has higher confidence for head classes (\figref{fig:unbalance}b). It is also an over-estimate of the true accuracy (\figref{fig:unbalance}c), especially for head classes. See further analysis in Appendix \ref{sec:extra-bias-effect}.

As an interesting side note, studies of human decision making and preference learning show a similar bias towards familiar classes. This effect is widely observed and has been connected to the availability heuristic studied by Tversky and Kahneman~\cite{Tversky1973AvailabilityAH}. 

A natural way to correct the familiarity bias would be to penalize high-frequency classes, either during training using a balanced loss, or post-training \cite{Kang2019DecouplingRA}. However, it would be a grave mistake to penalize all samples from rich classes, because confidence is sometimes justified, as in the case of "easy" prototypical examples of a class. Indeed, we show below that addressing the familiarity bias benefits from \textit{per-sample debiasing}, going beyond class-based debiasing.
To summarize, \textbf{model overconfidence is affected by class frequency. It can be estimated by observing the full vector of predictions to correct for overconfidence. Not all samples of a class should be penalized for belonging to a frequent class}.

Importantly, the familiarity effect caused by data imbalance has a crippling effect on model accuracy and on aggregating predictions from multiple modalities.
Several approaches attempted calibrating predictions of deep networks to remedy the above biases (see e.g. a survey in \cite{Guo2017OnCO}) and some became common practice. Unfortunately, the problem is still far from being solved.  

We propose to address both the \textit{data-imbalance} and the \textit{data paucity} learning challenges, using a single late-fusion architecture. 
We describe an easy-to-implement debiasing module that offsets the familiarity effect by learning to predict the magnitude of the bias for any given sample. It further improves learning at the tail by learning to fuse and balance information from visual and semantic modalities. It can easily be reduced to address long-tail learning with a single modality (vision only), where it improves over current baselines. 

~\newline
\noindent The paper has four main novel contributions: 
\noindent\newline\textbf{(1) A new late-fusion architecture} (\DRAGON{}) that learns to fuse predictions based on vision with predictions based class descriptors. 
\noindent\newline\textbf{(2) A module that rebalances class predictions across classes on a sample-by-sample basis.}
\noindent\newline\textbf{(3) New benchmarks }CUB-LT, SUN-LT, and AWA-LT for evaluating long-tail learning with textual class descriptors (LT-d).
    \DRAGON{} is SoTA on these datasets, and also on ImageNet-LT augmented with class descriptors and on existing two-level benchmarks.
\noindent\newline\textbf{(4) A new SoTA} on existing (vision-only) long-tail learning benchmarks: CIFAR-10, CIFAR-100, ImageNet-LT and Places365-LT.

\begin{figure*}[t]
    \centering
    \includegraphics[width=0.8\linewidth]{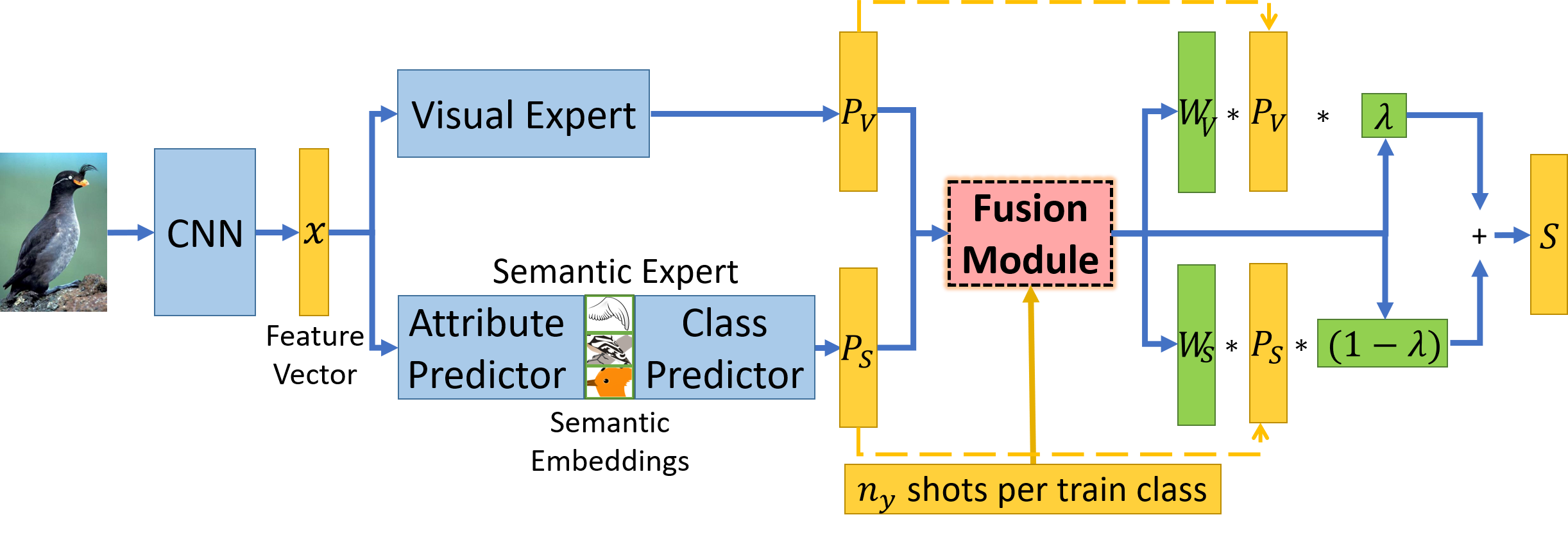}
    \caption{The \DRAGON{} architecture for long-tail learning with class-descriptors. The visual-expert and attribute-expert each outputs a prediction vector fed to a fusion module. The fusion module combines expert predictions and debias them. Blue, network components. Yellow, input to the fusion module. Green, the outputs of the fusion module. }
    \label{fig:DRAGON architecture}
\end{figure*}

\section{Related methods}
\label{sec:related}

\noindent\textbf{Long-tail learning:}
Learning with unbalanced data causes models to favor head classes ~\cite{buda2017systematic}. Previous efforts to address this effect can be viewed as either algorithmic or data-manipulations approaches. 

Algorithmic approaches encourage learning of tail classes using a non-uniform cost per misclassification function. A natural approach is to rescale the loss based on class frequency ~\cite{He_2009_IEEE}. \cite{Lin2017FocalLF} proposed to down-weigh the loss of well-classified examples, preventing easy negatives from dominating the loss. \cite{Ryou_2019_ICCV}  dynamically rescaled the cross-entropy loss based on the difficulty to classify a sample. \cite{cao2019learning} proposed a loss that encourages larger margins for rare classes.
\cite{Kang2019DecouplingRA} decoupled the learning procedure into representation learning and classification and studied four approaches. Among them, LWS $L_2$-normalizes the last-layer, since the weight magnitude correlates with class cardinality. The effect of this approach is similar to that presented in this paper, but here we apply recalibration dynamically on a sample-by-sample basis.

Data-manipulation approaches aim to flatten long-tail datasets to correct the bias towards majority classes. Popular techniques employ over-sampling of minority classes (more likely to overfit)\cite{Chawla2002SMOTESM,Han2005BorderlineSMOTEAN}, under-sampling the majority classes (wastes samples)\cite{Drummond2003C4}, or generating samples from the minority classes (can be costly to develop)\cite{Beery2019SyntheticEI}. 

Another approach is to transfer meta-level-knowledge from data-rich classes to data-poor classes.
\cite{Wang2017earningTM} gradually transfer hyperparameters from rich classes to poor classes by representing knowledge as trajectories in model space that capture the evolution of parameters with increasing training samples. \cite{openlongtailrecognition} first learns representations on the unbalanced data and then fine-tunes them using a class-balanced sampling and a memory module.

\noindent\textbf{Learning with class descriptors:}
Learning with class descriptors is usually applied to zero-shot learning (ZSL)~\cite{xianCVPR,DAP,LAGO}, where a classifier is trained to recognize (new) unseen classes based on their semantic description, which can include a natural-language textual description or predefined attributes. In several ZSL studies, attributes detected in a test image are matched with ground-truth attributes of each class, and several studies focused on this matching  ~\cite{DAP,LAGO,RelationNet,DEM,zhang_kernel,SYNC}.

A series of papers proposed to learn a shared representation of visual and text features (class-descriptors). As one example, 
\cite{REVISE} learns such a shared latent manifold using autoencoders and then minimizes the MMD loss between the two domains. Another recent line of work synthesizes feature vectors of unseen classes using generative models like VAE of GAN, and then use them in training a conventional classifier~\cite{PambalaGenerativeMW,xian_2018,CCGAN,CVAE2,CVAE1,ZhuGAN,Schnfeld2019GeneralizedZL,Xian2019FVAEGAND2AF}.  
The major baseline we compare our approach with is CADA-VAE~\cite{Schnfeld2019GeneralizedZL}, the current SoTA for Generalized FSL with class descriptors. CADA-VAE uses a variational autoencoder that aligns the distributions of image features and semantic (attribute) class embedding in a  shared latent space.
A recent work, ~\cite{Xian2019FVAEGAND2AF}, uses a mixture of VAEs and GANs.  
We could not directly compare with ~\cite{Xian2019FVAEGAND2AF} because their FSL protocol deviates from the standard benchmark of~\cite{Schnfeld2019GeneralizedZL,xianCVPR} by fine-tuning the CNN features. \textit{Without} fine-tuning, their reported metrics for GZSL are similar to CADA-VAE.

Some studies fused information from vision and \textit{per sample} descriptors (e.g., \cite{Zhang2016LearningDR}). This is outside the scope of this paper because it may require extensive labeling. 

\textbf{Generalized ZSL (GZSL) and Generalized FSL:}  GZSL extends ZSL to the scenario where the test data contains both seen and unseen classes~\cite{chao,xianCVPR,socher2013zero}. Recently, GZSL extended to Generalized Few-Shot-Learning with class descriptors (GFSL-d), where the unseen classes are augmented with a fixed number of few training samples~\cite{Schnfeld2019GeneralizedZL,REVISE}.
Namely, the distribution of samples across classes is a 2-level distribution, with \textit{many} ``head" classes and a smaller set of ``tail" classes all having the same (small) number of samples per class. Both GZSL and GFSL-d can be viewed as special cases of long-tail learning with class-descriptors, but with a \textit{short}-tailed unnatural distribution.

Most related GZSL approaches are \cite{Atzmon2018AdaptiveCS,socher2013zero,ZhangGZSL}.  They use a gating mechanism to weigh the decisions of seen-classes experts and a ZSL expert. The gating module is modeled as an out-of-distribution estimator. The current paper differs from their work by (1) The problem setup is different. Here, all samples are \textit{in-distribution} and the distribution of classes is smooth and long-tail with a much smaller number of head classes. 
(2) \DRAGON{} architecture first quantifies and corrects the (smooth) familiarity effect. Then it learns how to fuse the debiased decision of the two experts.

\begin{figure*}
    \begin{center}
    \includegraphics[width=0.8\linewidth]{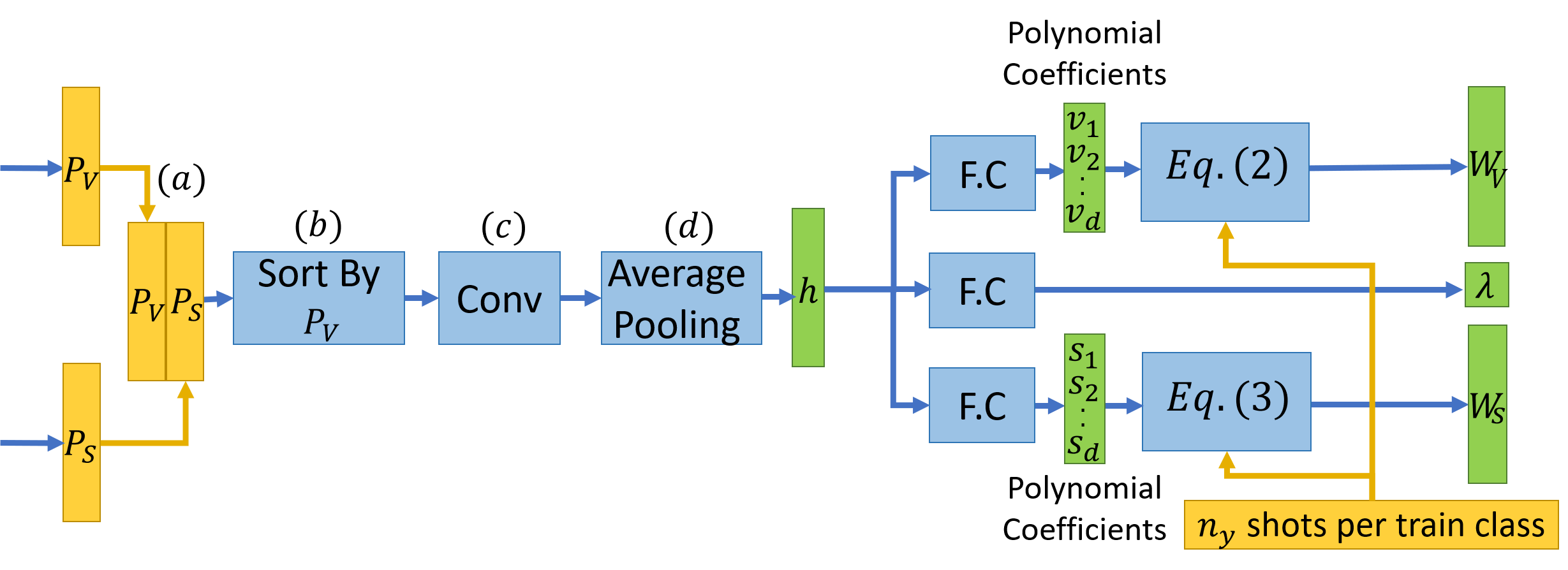}
    \end{center}
   \caption{Architecture of the fusion-module for long-tail learning with class-descriptors. In blue, network components. In yellow, inputs to the fusion-module and in green, activations or outputs of the fusion-module. The inputs $P_{V}$ denote the softmax prediction vector of the \textit{Visual Expert}, and $P_{S}$ that of the \textit{Semantic Expert}. The outputs $W_V$, $W_S$ and $\lambda$ are used in \eqref{eq_inference} for re-weighting the inputs. See \secref{sec:fusion_module} for more details.}
    \label{fig:fusion_module}
\label{fig:short}
\end{figure*}
 
\textbf{Early vs late fusion:}
When learning from multiple modalities, one often distinguishes between early and late fusion models~\cite{liu2018learn}. Early fusion models combine features from multiple modalities to form a joint representation. Late fusion methods combine decisions of per-modality models~\cite{Kahou2013CombiningMS,asvadi2018multimodal,pouyanfar2019multi}. Our approach addresses the long-tail setup, by leveraging the information in the familiarity bias to debias experts predictions.

\section{Long-tail learning with class descriptors}
\label{sec:setup}
We start with a formal definition of the problem of learning over unbalanced distributions with class descriptors. 

We are given a training set of $n$ labeled (image) samples: $\{(\x_1, \y_1),\ldots, (\x_n, \y_n)\}$, where each $\x_i$ is a feature vector and $\y_i$ is a label in $\{1,2, \dots k\}$. Samples are drawn from a distribution $\mathcal{D}=p(x,y)$ such that the marginal distribution over the classes $p(y)$ is strongly non uniform. For example, $p(y)$ may be exponential $p(y) \sim \exp(-ky)$. 

As a second supervision signal, each class $y$ is also accompanied with a  \textit{class-description} vector $\a_j$, $j=1,..,k$, in the form of semantic attributes~\cite{DAP} or natural-language embedding~\cite{Reed,ZhuGAN,socher2013zero}. For example, classes in CUB \cite{CUB} are annotated with attributes like \texttt{Head-color:red}.

At test time, a new set of $m$ test samples $\{\x_{n+1},\ldots,\x_{n+m}\}$ is given from the same distribution $\mathcal{D}$. We wish to predict their correct classes.

\section{Our approach}
\label{sec:approach}

Our approach is based on two observations: (1) Semantic descriptions of classes are easy to collect and can be very useful for tail (low-shot) classes, because they allow models to recognize classes even with few or no training samples~\cite{DAP,xianCVPR,LAGO} (and Appendix \ref{appendix:visual_vs_attribute}).
(2) The average prediction confidence over samples of a class is correlated with the number of training samples of that class (\figref{fig:unbalance}). 

Our architecture leverages these observations and learns to (1) Combine predictions of two expert classifiers: A conventional visual expert which is more accurate at head classes and a semantic expert which excels at tail classes; (2) Reweigh the scores of each class prediction, taking into account the number of training samples for that class.

\subsection{The architecture}
The \DRAGON{} architecture
\footnote{Dragons, like many distributions, have long-tails and are cool. For acronym lovers, \textsc{DRAGON} also stands for ``a moDulaR Approach for lonG-tail classificatiON''.} 
follows two design considerations: \textit{modularity} and \textit{low-complexity}. First, modularity;  \DRAGON{} allows to plug-in existing multi-modal experts, each trained for its own modality. Below we show experiments with language-based experts and a visual expert, but other modalities can be considered (e.g., mesh, depth, motion, or multi-spectral information).
Second, limiting the model to have a small number of parameters is important because tail classes only have few training samples and the model must perform well at the tail.

Our general architecture (\figref{fig:DRAGON architecture}) takes a late fusion approach. It consists of two experts modules: A visual expert and a semantic expert. Each expert outputs a prediction vector which is fed to a fusion module. The fusion module combines the expert predictions and learns to debias the familiarity effect, by weighing the experts and re-scaling their class predictions.

\subsection{A fusion-module}
\label{sec:fusion_module}
The fusion module takes as input the prediction vectors of two experts $p_V$, $p_S$ for a given image, and a vector containing the number of training samples per class. It has three outputs:  $\lambda\in(0,1)$ is a scalar to trade-off the visual expert against the semantic expert. 
$\FV \in (0,1)^k$ is a vector that weighs the predictions of the visual expert. Similarly, $\FA$ weighs the semantic expert predictions. Given these three outputs, a debiased score is computed for a class $y$:
\begin{eqnarray}
    S(y) &=& \lambda \FV(y) p_{\VE}(y) +  (1-\lambda)   \FA(y) p_{\AEx}(y).
    \label{eq_inference}
\end{eqnarray}
Figure \ref{fig:fusion_module} describes the architecture of the fusion-module. 
It has two main parts. The first part maps the prediction scores to a meaningful joint space, 
by first aligning the prediction of both classifiers, and then sorting according to confidence.

In more detail, the first part has four steps. (a) Stacking together the predictions of two experts to a $\Y \times 2$ vector. This makes the following convolution meaningful across the $2$ experts axis. (b) To make convolution meaningful also along the classes axis, and since classes are categorical, we reorder classes by their prediction score according to one of the experts. Section \ref{appendix:more-intuition} explains the rationale of this reordering. (c) Now that predictions are sorted, feed the sorted scores to a $N_{filters} \times 2 \times 2$ convolutional network. (d) Follow with an average-pooling layer (per class), yielding a $(\Y-1)$-dimensional vector $\h$. 

The goal of the second part is simply to predict a debiasing coefficient for each prediction, namely, learn a function from $n_y$ to $(0,1)^k$. We know from \figref{fig:unbalance} that the bias is inversely related to the number of samples. Aiming for a simple model, we train a polynomial regression that takes as input the number of samples and outputs a debiasing weight ($w_V(y)$). The coefficients of this polynomial $v_0,...,v_{d-1}$ are learned as a deep function over $\h$. Similarly for $f_S(y)$. 

More formally, let $n_y$ be the number of training samples of class $y$, and let $\bar{n}_y=n_y/\max_y n_y$ be the normalized counts, then we have
\begin{equation}
    \FV(y)=
    \sigma\Big[\sum_{j=0}^{d-1} v_j(\h)\bar{n}_y^j \Big] \,,\,\, 
    \FA(y)= 
    \sigma\Big[\sum_{j=0}^{d-1} s_j(\h)\bar{n}_y^j \Big], 
    \label{eq_polynom}
\end{equation}
where $d$ is the polynomial degree and $\sigma$ denotes a sigmoid that ensures that the resulting scale is in $[0,1]$.

Finally, the fusion module also predicts the trade-off scalar $\lambda$ to control the relative weight of the visual and semantic experts. This is achieved using a fully connected layer over $\h$. \secref{sec:ablation} analyzes the contribution of each component of the approach with an ablation study.

\subsection{Architecture design decision}
\label{appendix:more-intuition}
\DRAGON{} is designed with a small number of model parameters so it can improve predictions at the tail, where very few samples are available.

\textbf{Debiasing based on all expert predictions.}
To achieve the above goals, DRAGON implicitly learns how frequent is a class of a given sample. Namely, when the model receives a new test sample, it predicts if it is from a head class, a tail class, or somewhere between, and adjusts the confidence of the experts accordingly. To do this, it has been shown in the context of zero-shot learning that profile of confidence values are a good predictor if a sample comes from a seen or unseen class~\cite{Atzmon2018AdaptiveCS}. DRAGON generalizes this idea to long-tail distributions. To do this it takes as input all class predictions from both experts (\figref{fig:fusion_module}a). 

\textbf{Using order statistics over predictions.} To process expert prediction, we point out that order statistics over the prediction vector -- the maximum confidence, $2^{nd}$ max, etc\dots -- provides a strong signal about confidence calibration. Using the maximum of a vector is a very common operator in deep learning, known as max pooling. Here however, there is additional important information the gap between subsequent order statistics, like $\max - 2^{nd}\max$. As an intuitive example, a maximal prediction of 0.6 should be interpreted differently if the $2^{nd}$ max is $0.4$ or $0.1$. 

Order statistics can be easily computed by sorting the vector of predictions (\figref{fig:fusion_module}b). Sorting also increases the sample efficiency for learning, because later layers have each order statistic located at a fixed position in their input regardless of class. The function learned over order-statistics gaps is therefore shared across all classes.

The $2\times2$ convolution (\figref{fig:fusion_module}c) works well with the sorted expert predictions. Its filters capture two signals: (1) the confidence gaps between the two experts for each class; and (2) the confidence gaps between order-statistics for each expert alone.

\section{Experiments}
\label{sec:experimental setup overview}
\begin{figure*}[t!]
    \centering
    \includegraphics[width=0.24\linewidth]{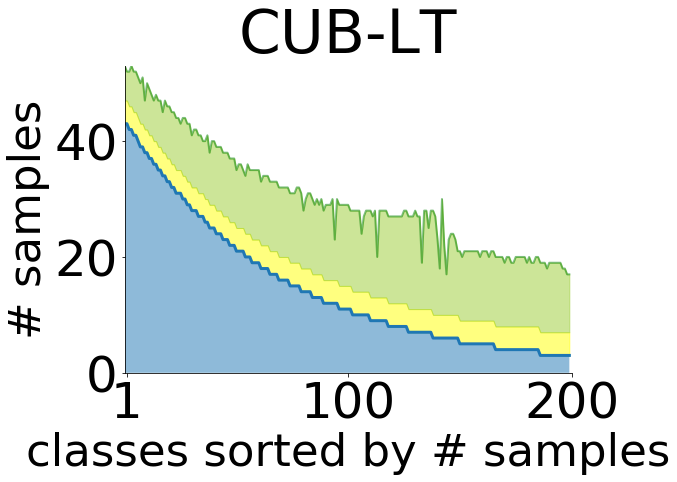}
    \includegraphics[width=0.24\linewidth]{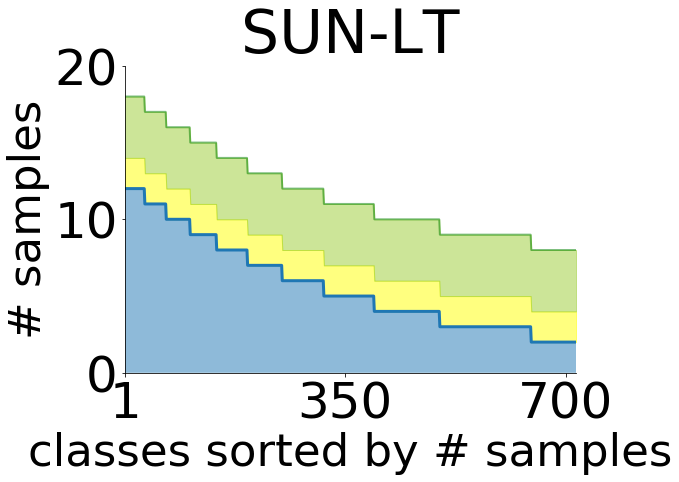}
    \includegraphics[width=0.24\linewidth]{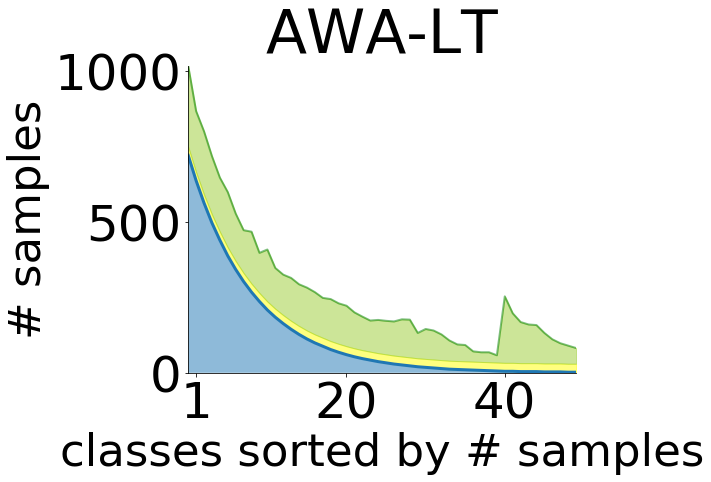}
    \includegraphics[width=0.24\linewidth]{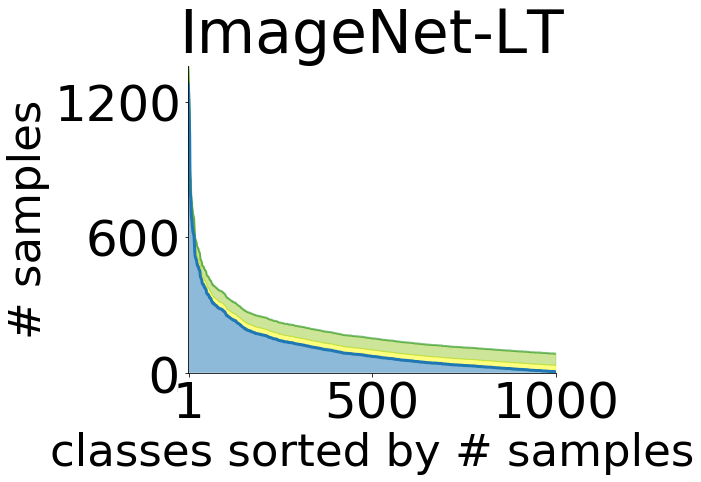}
    \caption{Long-tailed versions of CUB, SUN, AWA and ImageNet. Number of samples for the training, validation and test sets are shown respectively by blue, yellow and green. }%
    \label{fig:longtail}
\end{figure*}

We evaluate  \DRAGON{} in three unbalanced benchmark scenarios. (1) \textit{``\SmoothTail{}''}, the long-tailed distribution of classes decays smoothly (\figref{fig:longtail}) and each class is accompanied with textual class descriptors. (2) \textit{``\TwoLevel{}''}, the distribution has a step-shape as in~\cite{Schnfeld2019GeneralizedZL}; Most classes have many samples and the rest have few samples (\figref{fig:twostep}). 
(3) \textit{"Vision-only"}, a long-tail setup, as in \SmoothTail{}, but without class descriptors.

We compare DRAGON with SoTA approaches on standard benchmarks for each of these three scenarios. See Appendix \ref{sec:implemntation} for implementation details. 

Code available at \url{https://github.com/dvirsamuel/DRAGON}

\subsection{Overview of main results}
\textbf{For \SmoothTail{} distributed data}, we evaluated \DRAGON{} on four  benchmarks that we created from existing datasets. We added textual descriptors to ImageNet-LT \cite{openlongtailrecognition}, and generated long-tail versions of CUB, SUN and AWA. Dragon outperforms all baselines on various metrics. 

\textbf{For \TwoLevel{} distributed data}, \DRAGON{} surpasses the current SoTA \cite{Schnfeld2019GeneralizedZL}, tested using their experimental setup. 

\textbf{For Vision-only long-tail data}, we tested the calibration component of \DRAGON{} (without fusion). It achieves a new SoTA on ImageNet-LT \cite{openlongtailrecognition}, Places365-LT \cite{openlongtailrecognition}, Unbalanced CIFAR-10/100 \cite{cao2019learning} and comparable results on iNaturalist2018~\cite{Horn2017TheIC}.

\section{\SmoothTail{} distribution} 
\label{sec:smooth-tail}
\subsection{Datasets}
To evaluate %
long-tail learning with class descriptors, we created benchmark datasets in two ways. First, we created long-tail versions of existing learning-with-class-descriptors benchmarks. Second, we augmented existing long-tail benchmark (ImageNet-LT) with class descriptors. 

Specifically, we created new long-tail variants of the 3 main learning-with-class-descriptors benchmarks: CUB~\cite{CUB}, SUN~\cite{SUN} and AWA~\cite{DAP}, illustrated in \figref{fig:longtail}.
We ranked classes by the number of samples in each class after assigning tail classes to be consistent with those in the \TwoLevel{} benchmark \cite{xian_2018, Schnfeld2019GeneralizedZL} (See Appendix \ref{appendix:clarification_class_alignment} for more details). We then computed a frequency level for each class following an exponentially decaying function of the form $f(class) =$ $a b^{-rank(class)}$. $a$ and $b$ were selected such that the first class has the maximum number of samples, and the last class has 2 or 3 samples depending on the dataset. We then drew a random subset of samples from each class based on their assigned frequency $f(class)$. To create the validation set, we randomly drew a constant number of samples per class, while keeping an overall size of 20\% of the training set. 
See dataset statistics in Table \ref{table:train-splits}.

\begin{table}[t!]
    \centering
    \scalebox{1}{ 
    \setlength{\tabcolsep}{4pt} 
    \hskip-.5cm
    \begin{tabular}{l|c|c|c}
    & \textbf{CUB-LT} & \textbf{SUN-LT} & \textbf{AWA-LT}\\
    \midrule
    Train \# samples & 2,945 & 4,084 & 6,713    \\
    Val \# samples & 600 & 1,434 & 1,250 \\
    Test \# samples & 2,348 & 2,868 & 6,092  \\
    \hline
    \textbf{Train Set Properties} &&&    \\
    Max \# samples & 43 & 12 & 720   \\
    Min \# samples & 3 & 2 & 2 \\
    Mean \# samples & 14.725 & 5.696 & 123.460   \\
    Median \# samples & 11 & 5 & 35   \\
    \bottomrule
    \end{tabular} }
    \caption{Properties of CUB-LT, SUN-LT and AWA-LT.}
    \label{table:train-splits}
\end{table}

As a second type of benchmark, we used the existing long-tailed ImageNet~\cite{openlongtailrecognition} and augmented it with class descriptors. Specifically, we used the word2vec embeddings provided by \cite{SYNC}, which are widely used in the literature. 
Their word embeddings were created by training a skip-gram language model on a Wikipedia corpus to extract a 500-dimensional word vector for each class. See \cite{SYNC} Figure 4 for visualization.

Together, this process yielded the following datasets:
\newline1. \textbf{CUB-LT}, based on \cite{CUB},  consists of 2,945 training visual images of 200 bird species. Each species is described by 312 attributes (like tail-pattern:solid, wing-color:black). Classes have between 43 and 3 images per class. 
\newline2. \textbf{SUN-LT}, based on \cite{SUN}, consists of 4,084 training images, from 717 visual scene types and 102 attributes (like material:rock, function:eating, surface:glossy).
Classes have between 12 and 2 images per class.
\newline3. \textbf{AWA-LT}, based on \cite{DAP}, consists of 6,713 training images of 50 animal classes and 85 attributes (like texture:furry, or color:black).
Classes have between 720 and 2 images per class.
\newline4. \textbf{ImageNet-LT-d}, based on \cite{openlongtailrecognition}, consists of 115.8K images from 1000 categories with 1280 to 5 images per class. We use Word2Vec~\cite{Mikolov2013DistributedRO} class embeddings features provided by ~\cite{SYNC}, as textual descriptors.

\subsection{Training scheme} 
The familiarity effect is substantial in the validation and test data, but not in the training data, where models may actually become more confident on rare classes. We observed this effect in CUB-LT, SUN-LT, and AWA-LT. 
Since we wish to train the fusion module using data that exhibits the familiarity bias, we hold-out a subset of the training data and use it to simulate the response of experts to test samples. 
Note that in large-scale datasets, like ImageNet-LT-d, no hold-out set is needed and \DRAGON{} is trained on the training set. There, the familiarity bias is also present in the training data, as the models did not overfit the tail classes. Appendix \ref{sec:training-fusion-module} illustrates this effect in more detail.

\begin{table*}
\begin{subtable}{1\linewidth}
\sisetup{table-format=-1.2}   %
\centering
      \scalebox{1}{
\begin{tabular}{l|cc|cc|cc}
    {\textbf{(a)}} & \multicolumn{2}{c|}{\textbf{CUB-LT}} & \multicolumn{2}{c|}{\textbf{SUN-LT}} & \multicolumn{2}{c}{\textbf{AWA-LT}} \\
    {Method} & $Acc_{PC}$ & $Acc_{LT}$ & $Acc_{PC}$ & $Acc_{LT}$ & $Acc_{PC}$ & $Acc_{LT}$ \\
    \midrule
    \textbf{Vision-only} & & & & & & \\
    CE Loss* \textbf{(VE)} & 53.0 & 65.5 &  %
    33.7 & 40.0 &  %
    73.7 & 93.4 %
    \\
    Focal Loss~\cite{Lin2017FocalLF}*& 46.4 & 61.3 & 30.1 & 37.4 & 73.5 & 91.8 \\
    Anchor Loss~\cite{Ryou_2019_ICCV}*& 48.3 & 64.7 & 28.2 & 36.2 & 69.1 & 93.2  \\
    Range Loss~\cite{Zhang2017RangeLF}*& 48.5 & 65.3 & 27.9 & 36.0 & 68.9 & 93.5  \\
    LDAM Loss~\cite{cao2019learning}* & 50.1 & 64.1 & 29.8 & 36.4 & 69.1 & 93.5  \\
    CB LWS~\cite{Kang2019DecouplingRA}* & 53.1 & 65.7 & 33.9 & 40.2 & 73.4 & 93.6 \\
    \midrule
    \textbf{Multi-modal} & & & & & & \\
    LAGO~\cite{LAGO}* \textbf{(SE)} & 54.8 & 66.0 &  %
    18.2 & 18.3 & %
    74.0 & 93.0 %
    \\
    CADA-VAE~\cite{Schnfeld2019GeneralizedZL}* & 48.3 & 57.4 & 32.8 & 35.1 & 73.5 & 89.5 \\

\midrule
\textbf{Late Fusion} & & & & & & \\
    Mixture & 54.8 & 66.0  %
    & 34.0 &  40.3 %
    & 74.0 & 93.7 %
    \\
    \textbf{DRAGON (ours)} & \textbf{57.8} & \textbf{67.7} &%
    \textbf{34.8} & \textbf{40.4} & %
    \textbf{74.1}
    & \textbf{94.1}  %
    \\
    {\textbf{DRAGON + Bal'Loss (ours)}} & \textbf{60.1} & \textbf{66.5} &%
    \textbf{36.1} & 38.5 & %
    \textbf{76.2}
    & 92.2  %
    \\
    \bottomrule
    \end{tabular}
      }
\end{subtable}

\bigskip

\begin{subtable}{1\linewidth}
\sisetup{table-format=4.0} %
\centering

      \scalebox{1}{
            \begin{tabular}{l|c|ccc|c}
     {\textbf{(b) ImageNet-LT}}& {ResNet-10} & \multicolumn{4}{c}{{ResNeXt-50}} \\
    \hline
     {Method} & $Acc_{PC}$ & $Acc_{MS}$ & $Acc_{MED}$ & $Acc_{FS}$ & $Acc_{PC}$  \\
    \midrule
    
{\textbf{Vision-only}}&&&&&\\
    CE Loss* \textbf{(VE)} & 34.8 & 65.9 & 37.5 & 7.7 & 44.4 \\
    FSLwF~\cite{Gidaris2018DynamicFV} & 28.4 & - & - & - & -  \\
    Focal Loss~\cite{Lin2017FocalLF} & 30.5 & - & - & - & -  \\
    Range Loss~\cite{Zhang2017RangeLF} & 30.7 & - & - & - & -  \\
    OLTR~\cite{openlongtailrecognition}& 35.6 & - & - & - & 37.7 \\
    CB LWS~\cite{Kang2019DecouplingRA} & 41.4 & 60.2 & 47.2 & 30.3 & 49.9 \\
\midrule
{\textbf{Multi-modal}}&&&&&\\
DEM~\cite{DEM}* \textbf{(SE)} & 18.1 & 16.5 & 13.0 & \textbf{50.8} & 19.5
    \\
CADA-VAE~\cite{Schnfeld2019GeneralizedZL}* & 42.1 & 57.4 & 43.7 & 27.0 & 49.3
\\

\midrule
\textbf{Late Fusion} & & & & & \\
    Mixture & 40.7 & 63.8 & 36.3 & 23.9 & 45.1 \\
    Sharma et al.~\cite{sharma2020long} & 39.2 & - & - & - & - \\
    {\textbf{DRAGON (ours)}} & \textbf{43.1} & \textbf{66.0} & \textbf{38.3} & 47.6 & \textbf{51.2}
    \\
    {\textbf{DRAGON + Bal'Loss (ours)}} & \textbf{46.5} & 62.0 & \textbf{47.4} & 50.2 & \textbf{53.5}
    \\
    \bottomrule
    \end{tabular}
      }
     \end{subtable}
\caption{\textbf{Smooth-tail distribution:} Rows with * denote results reproduced by us. The rest were taken from  \cite{Kang2019DecouplingRA, openlongtailrecognition}. {VE} and {SE} refer to the \textit{visual-expert} and \textit{semantic-expert} that were used to train \DRAGON{}. Bal. refers to training \DRAGON{} with a balanced loss. \textbf{(a)}  Comparing \DRAGON{} with baselines on the long-tailed benchmark datasets. We report Per-Class Accuracy $Acc_{PC}$ and Long-Tailed Accuracy $Acc_{LT}$. \textbf{(b)} Comparing DRAGON with baselines on the long-tailed ImageNet with word embeddings. 
}

\label{fig:longtail-bench}
\end{table*}

\subsection{Baselines and variants}
We compared \DRAGON{} with long-tail learning and unbalanced data approaches: 
\textbf{Focal Loss~\cite{Lin2017FocalLF}}, \textbf{Anchor Loss~\cite{Ryou_2019_ICCV}},
\textbf{Range Loss~\cite{Zhang2017RangeLF}}, %
and \textbf{LDAM Loss~\cite{cao2019learning}} are loss manipulation approaches for long-tail distributions.
\textbf{FSLwF~\cite{Gidaris2018DynamicFV}}, \textbf{OLTR \cite{openlongtailrecognition}} and \textbf{Classifier-Balancing (CB) \cite{Kang2019DecouplingRA}} are algorithmic approaches in the long-tail learning benchmarks.
\textbf{Mixture} and \textbf{Class Balanced Experts~\cite{sharma2020long}} are late fusion approaches.
\textbf{Mixture} resembles mixture-of-experts (MoE) \cite{MoE_hinton} without EM optimization. It fuses the raw outputs of the two experts by a gating module. As with standard MoE models, the gating module is trained with visual-features as inputs.

For CUB-LT, SUN-LT, and AWA-LT, the visual expert was a linear layer over a pre-trained ResNet from~\cite{xianCVPR,xian_awa2}, which we trained with a balanced xent loss (CE Loss). The semantic expert was LAGO~\cite{LAGO}. For ImageNet-LT-d, we followed~\cite{Kang2019DecouplingRA} and set the visual expert to be ResNet-10 or ResNeXt-50. The semantic expert was DEM~\cite{DEM}.

\subsection{Evaluation metrics and protocol}

\textbf{Evaluation Protocol:}
The experiments for CUB-LT, SUN-LT, and AWA-LT follow the standard protocols set by \cite{Schnfeld2019GeneralizedZL,xianCVPR,xian_awa2}, including their ResNet-101 features.
Their split ensures that \textbf{none of the test classes appear in the training data used to train the ResNet-101 model}. 
For ImageNet-LT-d we used the protocols in ~\cite{openlongtailrecognition,Kang2019DecouplingRA} with the pre-trained ResNet-10 and ResNeXt-50 provided by ~\cite{Kang2019DecouplingRA}. 

\begin{table}
\begin{center}
    \scalebox{1}{
    \setlength{\tabcolsep}{1pt} %
    \hskip-0.2cm
    \begin{tabular}{lcc}
    \textbf{CUB-LT} & \textbf{\(Acc_{PC}\)} &
    \textbf{\(Acc_{LT}\)} \\
    \midrule
    Max. & 57.3 & 70.8    \\
    Avg. & 57.0 & 70.0   \\
    Product & 56.9 & 70.3   \\
    Mixture & 53.7 & 66.5   \\
    \midrule
    \textbf{DRAGON (ours)} & \textbf{60.0} & \textbf{70.9} \\
    \bottomrule
    \end{tabular}
    }
\end{center}
\caption{Comparing \DRAGON{} against common late-fusion approaches on the validation set of CUB-LT.}
\label{fusion-bench}
\end{table}

\textbf{Evaluation metrics:}
We evaluated \DRAGON{} on the \SmoothTail{} benchmark with the following metrics: 
\newline\textbf{(a) Per-Class Accuracy} ($Acc_{PC}$): Balanced accuracy metric that uniformly averages the accuracy of each class  $\frac{1}{k}\sum_{y=1}^k{Acc(y)}$, where $Acc(y)$ is the accuracy of class $y$.  
\newline\textbf{(b) Long-Tailed Accuracy} ($Acc_{LT}$): Test accuracy, where the distribution over test classes is long-tailed like the training distribution. This is expected to be the typical case in real-world scenarios. See Appendix \ref{appendix:long-tail-eval} for more details.
\newline\textbf{(c) Many-Shot}, \textbf{Medium-shot} and \textbf{Few-Shot accuracies}: For ImageNet-LT-d, we follow \cite{openlongtailrecognition} and report accuracy for: $Acc_{MS}$ ($>$\!100 training images), $Acc_{MED}$ (20-100 images) and $Acc_{FS}$ ($<$ 20 images).

\subsection{Results with smooth-tail distribution}
Table \ref{fig:longtail-bench} (a) provides the test accuracy for three long-tail benchmark datasets and compares \DRAGON{} to baselines and individual components of the \DRAGON{} model. \DRAGON{} achieves higher accuracy compared with all competing methods, both with respect to class-balanced accuracy ($Acc_{PC}$) and to test-distribution accuracy ($Acc_{LT}$).
Improving $Acc_{LT}$ indicates that \DRAGON{} effectively classifies head classes, which are heavily weighted in $Acc_{LT}$. At the same time, improving $Acc_{PC}$ indicates that \DRAGON{} also effectively classifies tail classes, which are up-weighted in $Acc_{PC}$. 
\newline\indent Table \ref{fig:longtail-bench} (b) provides the $Acc_{PC}$ accuracy for ImageNet-LT-d. We can directly see the benefit of fusion information between modalities - the visual expert excels on many-shot classes, $Acc_{MS}$, while the semantic expert excels only on few-show classes, $Acc_{FS}$. \DRAGON{} recalibrate and fuse both experts to excel in all classes.

We further trained \DRAGON{} with a balanced cross-entropy loss (Bal.). This strategy has a synergistic effect with \DRAGON{} (last row in Table \ref{fig:longtail-bench}): It improves tail accuracy $Acc_{PC}$ for all benchmarks while only marginally hurting head accuracy $Acc{LT}$.
 
Table \ref{fusion-bench}  compares \DRAGON{} against common late fusion strategies, on the validation set of CUB-LT:  \textit{AVG} (averaging expert predictions), \textit{Max} (taking the largest prediction), \textit{Product} (multiplying expert predictions) and \textit{Mixture}.
We show that those approaches, which late fuse predictions of the two experts, are usually better at head classes ($Acc_{LT}$) while giving less accurate results for tail classes ($Acc_{PC}$). \DRAGON{} achieves better results on both metrics because it also calibrates expert predictions. In the ablation study (Table \ref{architecture-ablation-bench}) we compare our fusion module to more ablated fusion components.

\section{\TwoLevel{} (GFSL-d) benchmark}
\label{sec:two-level}

\begin{figure}[t!]
    \centering
    \includegraphics[width=0.31\linewidth]{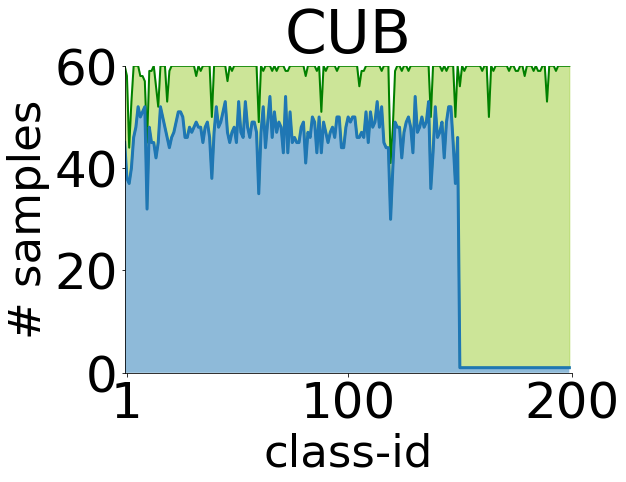}
    \includegraphics[width=0.31\linewidth]{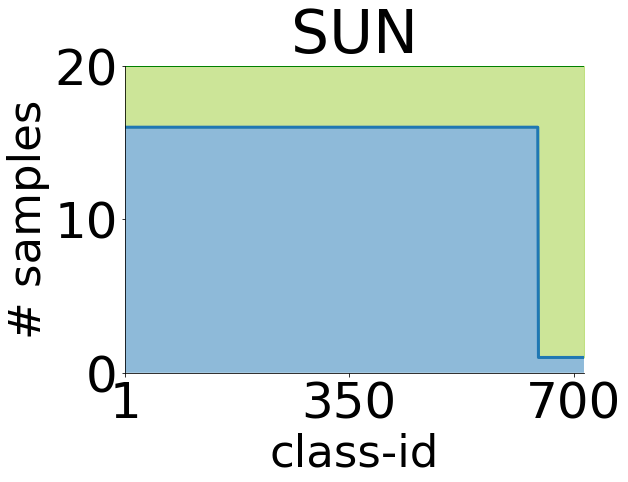}
    \includegraphics[width=0.31\linewidth]{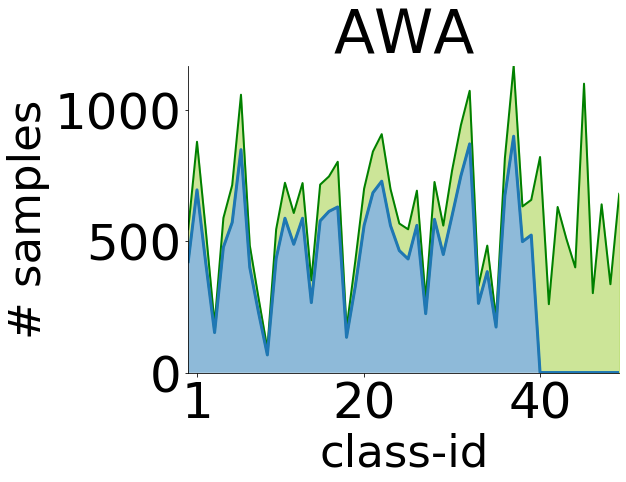}
    \caption{Two-level variants of CUB, SUN and AWA as in ~\cite{Schnfeld2019GeneralizedZL}. Blue: training set, green: test set. 
    }
    \label{fig:twostep}
\end{figure}

We follow the protocol of ~\cite{Schnfeld2019GeneralizedZL} on the original CUB, SUN, and AWA (\figref{fig:twostep}), to compare \DRAGON{} in a \TwoLevel{} setting.

For those datasets, many-shot classes are kept as in the original train-set, while few-shot classes have an increasing number of shots: 1,2,5,10 and 20 (in SUN up to 10 shots).

\subsection{Baselines and variants}
We compared \DRAGON{} with \textbf{LDAM Loss \cite{cao2019learning}} and with SoTA multi-modal GFSL-d  approaches: 
\textbf{ReViSE~\cite{REVISE}}, \textbf{CA-VAE~\cite{Schnfeld2019GeneralizedZL}},
\textbf{DA-VAE~\cite{Schnfeld2019GeneralizedZL}} and
\textbf{CADA-VAE~\cite{Schnfeld2019GeneralizedZL}}. Their results were obtained from the authors of \cite{Schnfeld2019GeneralizedZL}, while LDAM results were reproduced by us.

The visual expert was a linear layer over a pre-trained ResNet from~\cite{xianCVPR,xian_awa2}, which we trained with a balanced cross-entropy loss. The semantic expert was LAGO~\cite{LAGO}.

 \begin{table*}[t!]

    \begin{center}
    \scalebox{1}{
    \setlength{\tabcolsep}{4pt} %
    \begin{tabular}{lccccc|cccc|ccccc}
     Two-Level & 
    \multicolumn{5}{c|}{\textbf{CUB}} & \multicolumn{4}{c|}{\textbf{SUN}} & \multicolumn{5}{c}{\textbf{AWA}}\\
    \textbf{\# shots} & \textbf{1} & \textbf{2} & \textbf{5}  & \textbf{10}  & \textbf{20} &
    \textbf{1} & \textbf{2} & \textbf{5}  & \textbf{10}  & 
    \textbf{1} & \textbf{2} & \textbf{5}  & \textbf{10} & \textbf{20} \\
\midrule
    LDAM~\cite{cao2019learning}* & 2.4 & 10.9 & 36.0 & 52.2 & 61.5 &%
    4.3 & 11.5 & 26.6 & 37.0 & %
    12.4 & 24.8 & 41.1 & 57.0 & 68.6 %
    \\
    \midrule
    REVISE~\cite{REVISE} & 36.3 & 41.1 & 44.6 & 50.9 & - & %
    27.4 & 33.4 & 37.4 & 40.8 & %
    56.1 & 60.3 & 64.1 & 67.8 & - %
    \\ 
    CA-VAE~\cite{Schnfeld2019GeneralizedZL} & 50.6 & 54.4 & 59.6 & 62.2 & - & %
    37.8 & 41.4 & 44.2 & 45.8 &  %
    64.0 & 71.3 & 76.6 & 79.0  & -  %
    \\
    DA-VAE~\cite{Schnfeld2019GeneralizedZL} & 49.2 & 54.6 & 58.8 & 60.8 & - & %
    37.8 & 40.8 & 43.6 & 45.1 &  %
    68.0 & 73.0 & 75.6 & 76.8 & -   %
    \\
    CADA-VAE~\cite{Schnfeld2019GeneralizedZL} & 55.2 & 59.2 & 63.0 & 64.9 & 66.0 & %
    40.6 & 43.0 & 46.0 & 47.6 & %
    \textbf{69.6} & \textbf{73.7} & \textbf{78.1} & 80.2 & 80.9  %
    \\ 
    \midrule
    CE Loss* \textbf{(VE)} & 1.2 & 6.9 & 30.2 & 50.2 & 60.9 & %
    1.8 & 8.9 & 25.1 & 38.3 & %
    11.0 & 20.0 & 47.8 & 69.9 & 73.9 %
    \\ 
    LAGO* \textbf{(SE)} & 23.0 & 33.2 & 49.0 & 58.6 & 64.8 & %
    19.5 & 23.2 & 25.6 & 27.8 & %
    20.2 & 33.0 & 59.0 & 68.7  & 75.8  %
    \\
    \textbf{DRAGON (ours)}  & \textbf{55.3} & \textbf{59.2} & \textbf{63.5} & \textbf{67.8} & \textbf{69.9} &%
    \textbf{41.0} & \textbf{43.8} & \textbf{46.7} & \textbf{48.2} & %
    67.1 & 69.1 & 76.7 & \textbf{81.9} & \textbf{83.3} %
    \\ 
    \bottomrule
    \end{tabular}
    }
    \end{center}
    \vspace{-12pt}
  \caption{\textbf{\TwoLevel{} distributions:} Comparing  \DRAGON{} on \TwoLevel{} CUB, SUN and AWA with SoTA GFSL models and baselines and with increasing number of few-shot training samples. Values denote  the Harmonic mean $Acc_{H}$.
  VE and SE refer to the \textit{visual-expert} and \textit{semantic-expert} that were used to train DRAGON.
  }  
\label{two-level-bench}

\end{table*}

\subsection{Evaluation metrics}
Following  ~\cite{Schnfeld2019GeneralizedZL}, we evaluated the \TwoLevel{} benchmark with the Harmonic mean metric  ($Acc_{H}$): It quantifies the overall performance of head and tail classes, by $Acc_{H} = 2(Acc_{ms}Acc_{fs})/(Acc_{ms} + Acc_{fs})$. Where, $Acc_{ms}$ is the per-class accuracy over many-shot classes and $Acc_{fs}$ is the per-class accuracy over few-shot classes.

\subsection{Results with two-level distribution}

Table \ref{two-level-bench} %
compares \DRAGON{} with SoTA baselines on the \TwoLevel{} setup. Our model wins in CUB and SUN on all shots but loses on AWA for fewer than 10 samples. Furthermore, \DRAGON{}  gains better results when the number of shots increases in contrast to complex generative models like CADA-VAE~\cite{Schnfeld2019GeneralizedZL}. Appendix \ref{appendix:ms-fs-two-level}
provides results for $Acc_{fs}$ and $Acc_{ms}$ with 1,2,5,10,20 shots.

\section{Vision-only long-tail learning}
\label{sec:extension}
The approach presented in this paper focuses on learning from two modalities, vision and language. To understand the effect of re-calibrating we now study a simpler variant of \DRAGON{} that can be applied to the more common vision-only long-tail learning.
We name it sm\DRAGON{} for \textit{single-modality-\DRAGON{}}, and show that it achieves new state-of-the-art results, compared to uni-modals baselines, on ImageNet-LT, Places365-LT, CIFAR-10 and CIFAR-100. On iNaturalist sm\DRAGON{} is comparable to SoTA.
To adapt to single-modality, we train sm\DRAGON{} only on the predictions of the visual-expert.
It outputs a single set of coefficients $\{\FV(y)\}_{y\in\Y}$ to rescale the predictions of the visual expert, instead of two sets of coefficients. Subsequently, \eqref{eq_inference} reduces to 
$    S(y) = \FV(y) p_{\VE}(y)$.

In other words, sm\DRAGON{} is a simplified version of DRAGON that is trained on the predictions of the \textbf{visual-expert only} (no class descriptors being used). 
sm\DRAGON{} takes the predictions of a freezed visual-expert and rescale it by learning a single set of polynomial coefficients. During inference, sm\DRAGON{} balances the visual-expert predictions in a sample-by-sample basis.

\begin{table*}[t!]
    \begin{center}
      \scalebox{1}{
    \setlength{\tabcolsep}{4pt} %
\hskip-.3cm
    \begin{tabular}{l|cc|cc|cc|cc} 
 \hline
    Vision-Only &
    \multicolumn{4}{c|}{\textbf{Unb. CIFAR-10}} & \multicolumn{4}{c}{\textbf{Unb. CIFAR-100}} \\
 \hline
    Imbalance Type &
    \multicolumn{2}{c|}{long-tail} & \multicolumn{2}{c|}{two-level} & \multicolumn{2}{c|}{long-tail} & 
    \multicolumn{2}{c}{two-level}\\
 \hline
 Imbalance Ratio & 100 & 10 & 100 & 10 & 100 & 10 & 100 & 10 \\ [0.5ex] 
  \hline
~\cite{Cui2019ClassBalancedLB} ReSample &  29.45 &  13.21 &  38.14 &  15.41 &  66.56 &  44.94 &  66.23 &  46.92 \\
 ~\cite{Cui2019ClassBalancedLB} ReWeight &  27.63 &  13.46 &  38.06 &  16.20 &  66.01 &  42.88 &  78.69 &  47.52 \\
 ~\cite{Cui2019ClassBalancedLB} Focal &  25.43 &  12.90 &  39.73 &  16.54 &  63.98 &  42.01 &  80.24 &  49.98 \\
 \hline
 CE~\cite{cao2019learning}&   29.64 &  13.61 &  36.70 &  17.50 &  61.68 &  44.30 &  61.45 &  45.37 \\
  CE* \textbf{(VE)} &  29.81 &  13.12 &  36.61 &  17.78 &  61.72 &  43.77 &  61.59 &  45.75 \\
 Focal~\cite{Lin2017FocalLF} &  29.62 &  13.34 &  36.09 &  16.36 &  61.59 &  44.22 &  61.43 &  46.54 \\
 LDAM~\cite{cao2019learning} &  26.65 &  13.04 &  33.42 &  15.00 &  60.40 &  43.09 &  60.42 &  43.73 \\
 \hline
 \textbf{sm\DRAGON{} (ours)} &  \textbf{22.08} &  \textbf{12.17} &  \textbf{27.10} &  \textbf{12.38} &  \textbf{58.01} &  \textbf{42.22} &  \textbf{54.43} &  \textbf{40.97} \\
\bottomrule
\end{tabular}}
    \end{center}
      \caption{\textbf{Vision-only long-tail}. Error rate of ResNet32 on unbalanced CIFAR-10 and CIFAR-100~\cite{cao2019learning}, comparing sm\DRAGON{} and SoTA techniques. sm\DRAGON{} was trained over predictions of the cross-entropy model (CE*). Reported values are the top-1 validation error. Asterisks * denote results reproduced using published code.}
  \label{cifar-bench}
\end{table*}

 \begin{table*}[t!]
    \begin{center}
      \scalebox{1}{
    \setlength{\tabcolsep}{4pt} %
\hskip-.3cm
 \begin{tabular}{l|cc|cc|cc|cc} 
 \hline
    Vision-Only  &
    \multicolumn{4}{c|}{\textbf{Unb. CIFAR-10}} & \multicolumn{4}{c}{\textbf{Unb. CIFAR-100}} \\
 \hline
    Imbalance Type &
    \multicolumn{2}{c|}{long-tail} & \multicolumn{2}{c|}{two-level} & \multicolumn{2}{c|}{long-tail} & 
    \multicolumn{2}{c}{two-level}\\
 \hline
 Imbalance Ratio & 100 & 10 & 100 & 10 & 100 & 10 & 100 & 10 \\ [0.5ex] 
  \hline
   CE-DRW* \textbf{(VE1)} &  24.73 &  13.52 &  28.65 &  13.90 &  59.23 &  42.19 &  58.93 &  45.00 \\
  M-DRW~\cite{Wang2018AdditiveMS} &  24.94 &  13.57 &  26.67 &  13.17 &  59.49 &  43.49 &  58.91 &  44.72 \\
 LDAM-DRW~\cite{cao2019learning} &  22.97 &  11.84 &  23.08 &  12.19 &  57.96 &  41.29 &  54.64 &  40.54 \\
  LDAM-DRW* ~\cite{cao2019learning} \textbf{(VE2)} &  22.96 &  11.84 &  23.41 &  12.20 &  57.89 &  41.61 &  54.65 &  43.48 \\
 
  \hline
  \textbf{VE1 + sm\DRAGON{}} &  \textbf{20.37} & 12.06 &  21.54 &  \textbf{11.94} & \textbf{56.50} &  42.11 & \textbf{53.32} &  40.66 \\
 \textbf{VE2 + sm\DRAGON{}} &  21.22 &  \textbf{11.84} &  \textbf{20.64} &  12.37 &  56.70 &  \textbf{41.23} &  54.07 &  \textbf{40.35} \\
\bottomrule
\end{tabular}}
    \end{center}
        \caption{\textbf{Vision-only long-tail:} sm\DRAGON{} was trained on top models trained with DRW (VE1) or LDAM-DRW (VE2) ~\cite{cao2019learning}. Similar to Table \ref{cifar-bench} except that all models were trained with DRW schedule \cite{cao2019learning}.
   Reported values are top-1 validation error. Asterisks * denote results that we reproduced using code published by the authors of~\cite{cao2019learning}.}
  \label{cifar-bench-drw}
\end{table*}

\begin{table*}[t!]
    \centering
      \scalebox{1}{ 
    \setlength{\tabcolsep}{1pt} 
\begin{tabular}{l|c|c|c}
    {Vision-Only} & \textbf{Places365-LT} & \multicolumn{2}{c}{\textbf{ImageNet-LT}} \\
    {} & {ResNet-50} & {ResNet-10} & {ResNeXt-50} \\
    \midrule
    CE Loss* \textbf{(VE)} & 30.2  & 34.8 & 44.4 \\
    Bal' Loss & 32.4 & 33.1 & - \\
    Lifted Loss~\cite{Song2016DeepML} & 35.2 & 30.8 & - \\
    Focal Loss~\cite{Lin2017FocalLF} & 34.6 & 30.5 & - \\
    Range Loss~\cite{Zhang2017RangeLF} & 35.1 & 30.7 & - \\
    FSLwF~\cite{Gidaris2018DynamicFV} & 34.9  & 28.4 & -\\
    OLTR~\cite{openlongtailrecognition} & 35.9 & 34.1 & 37.7  \\
    CB $\tau-$norm~\cite{Kang2019DecouplingRA} & 37.9 & 40.6 & 49.4 \\
    CB LWS~\cite{Kang2019DecouplingRA} & 37.6 & 41.4 & 49.9  \\
\midrule
    \textbf{smDRAGON (ours)} & \textbf{38.1} & \textbf{42.0} & \textbf{50.1} 
    \\
    \bottomrule
    \end{tabular}
 \,\,\,\,\,\,\,\,
\begin{tabular}{l|c}
    {Vision-Only} & \textbf{iNaturalist}\\
    {} & {ResNet-50} \\
    \midrule
    \cite{Cui2019ClassBalancedLB} Focal & 61.1   \\
    LDAM~\cite{cao2019learning} & 64.6  \\
    LDAM-DRW~\cite{cao2019learning} & 68.0  \\
    CB $\tau-$norm~\cite{Kang2019DecouplingRA} & 69.3 \\
    CB LWS~\cite{Kang2019DecouplingRA} & 69.5   \\
\midrule
    \textbf{smDRAGON (ours)} & 69.1
    \\
    \bottomrule
    \end{tabular}
    }
    \caption{ \textbf{ Vision-only long-tail:} Baseline results where copied directly from ~\cite{cao2019learning} and ~\cite{Kang2019DecouplingRA}. \textbf{Left:} sm\DRAGON{} achieves better $Acc_{PC}$ on Places365-LT and ImageNet-LT. \textbf{Right:} Comparing sm\DRAGON{} on long-tailed iNatrualist. sm\DRAGON{} achieve comparable results compared to SoTA baselines. 
  }
  \vspace{-5pt}
  \label{fig:single-modality-bench}
\end{table*}

Tables \ref{cifar-bench} and \ref{cifar-bench-drw} compare sm\DRAGON{} against approaches in the unbalanced CIFAR-10 and CIFAR-100 benchmarks, as presented in~\cite{cao2019learning}:
\textbf{CE Loss}, \textbf{Resample~\cite{Cui2019ClassBalancedLB}}, \textbf{Reweight~\cite{Cui2019ClassBalancedLB}}, \textbf{Focal~\cite{Cui2019ClassBalancedLB}} and \textbf{LDAM Loss~\cite{cao2019learning}}. 
\textit{DRW}~\cite{cao2019learning} denotes models that were trained with the training schedule proposed by~\cite{cao2019learning}.  

Table \ref{fig:single-modality-bench} compares sm\DRAGON{} with popular baselines and recent long-tail learning approaches in the ImageNet-LT and Places-LT benchmarks. Those are the same baselines as in the \SmoothTail{} setup (\secref{sec:smooth-tail}).

\begin{figure}[t]
  \includegraphics[trim=0 50 0 0, clip,width=\linewidth] {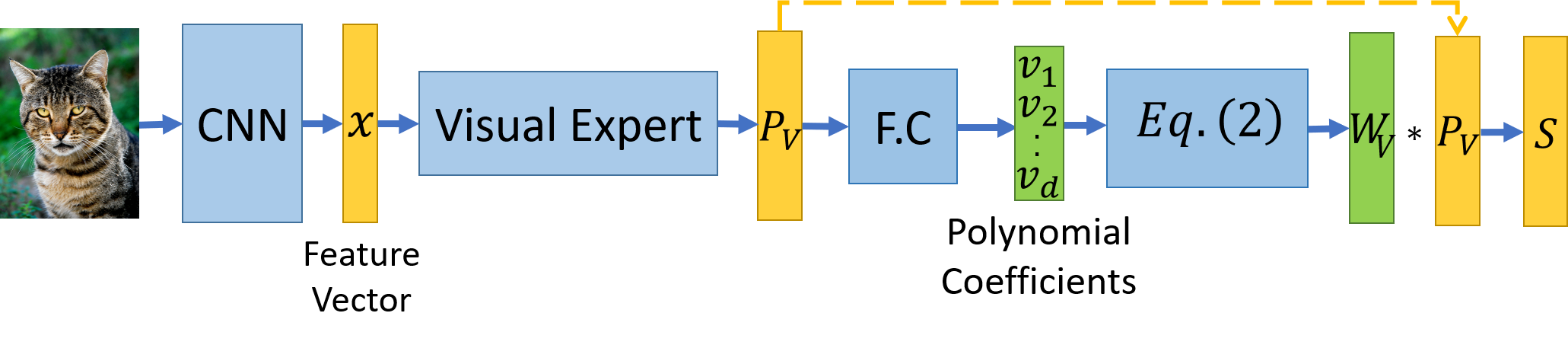}
    \caption{Architecture of vision-only sm\DRAGON{}.}
    \label{fig:smDragon_architecture}
    \vspace{-10pt}
\end{figure}

The results demonstrate that (1) sm\DRAGON{} outperforms all baselines and (2) combining sm\DRAGON{} with SoTA approaches (LDAM or DRW) has a synergistic effect.

Table \ref{tab:smDragon_test} compares \DRAGON{} with sm\DRAGON{} on CUB-LT. It shows that fusing information between modalities (third row) gives better results than re-scaling expert predictions alone (first and second rows). Finally, on iNaturalist 2018, sm\DRAGON{} is comparable to SoTA, reaching 69.1\% compared to 69.5\% (CB-LWS \cite{Kang2019DecouplingRA}).

\begin{table*}[t!]
    \begin{center}
    \begin{sc}
    \scalebox{1}{
    \setlength{\tabcolsep}{1pt} %
    \hskip-0.2cm
    \begin{tabular}{lccc}
    {\textbf{}} & \textbf{\(Acc_{PC}\)} &
    \textbf{\(Acc_{LT}\)} &
    \textbf{\(\#params\)} \\
    \midrule
    F.C. & 54.0 & 67.1 & 403    \\
    F.C. $\&$ $1/n_y$ rescale & 56.7 & 60.3 & 403    \\
    F.C. $\&$ non-parametric rescale & 58.2 & 68.0 & 81,406    \\
    Conv. $\&$ non-parametric rescale & 58.7 & 68.2 & 40,612    \\
    Conv. $\&$ single parametric rescale & 59.0 & 67.5 & 613    \\
    \textbf{DRAGON (ours)} & \textbf{60.0} & \textbf{70.9}  & \textbf{1,015}  \\
    \bottomrule
    \end{tabular}
    }
    \end{sc}
    \end{center}
  \caption{Ablation study, comparing different fusion and re-scaling approaches.
  The results show the contribution of the convolutional backbone and the re-scaling method for the two experts (validation set, CUB-LT).}
  \label{architecture-ablation-bench}
  
\end{table*}

\begin{table}[t!]
    \centering
    \scalebox{0.8}{
    \setlength{\tabcolsep}{1pt}
    \hskip-.5cm
    \begin{tabular}{l|c|c|c|c}
    {} & \textbf{Places365-LT} & \multicolumn{2}{c}{\textbf{ImageNet-LT}} & \textbf{CIFAR100-LT}\\
    {} & {ResNet-50} & {ResNet-10} & {ResNeXt-50} & {ResNet-32}\\
    \midrule
    CE Loss \textbf{(VE)} & 30.2  & 34.8 & 44.4 & 38.3\\
    Per-class & 36.9 & 40.0 & 49.2 & 40.5\\
    Per-sample & \textbf{38.1} & \textbf{42.0} & \textbf{50.1} & \textbf{42.0} \\
    \bottomrule
    \end{tabular}}
    \caption{Ablation of per-sample weighting on vision-only benchmarks. VE refers to \textit{visual-expert}.}
    \label{tab:sample-basis}
\end{table}

\section{Ablation study}
\label{sec:ablation}
To understand the contribution of individual components of \DRAGON{}, we carried ablation experiments. We report results on the validation set, which were consistent with the test set (Appendix \ref{appendix:ablation-test-set}).

\textbf{Fusion-Module Architecture:}
Table \ref{architecture-ablation-bench} compares the performance of various components of the fusion-module on CUB-LT.
(1) \textit{F.C.:} predicts $\lambda$ using a fully-connected layer over $\vec{p}_{\VE}, \vec{p}_{\AEx}$, no re-balancing ($\FV(y)=\FA(y)=1, \forall y$). 
(2) \textit{F.C. $\&$ $1/n_y$ rescale:} learns $\lambda$ as in \textit{F.C.}, rescales experts predictions by $n_y$. (3) \textit{F.C. $\&$ non-parametric rescale:} learns $\lambda$ as \textit{F.C.} and rescales both experts predictions by a learned non-parametric weight for each class instead of a polynomial.
(4) \textit{Conv. $\&$ non-parametric rescale:} like (3), then applies sorting and convolution (\secref{sec:fusion_module}).
(5) \textit{Conv. $\&$ single parametric rescale} replaces the non-parametric re-scaling weights by a single polynomial of parametrized weights. (6) \DRAGON{} is our full approach  described in \secref{sec:approach}.
The comparison shows that rescaling expert predictions significantly improves $Acc_{PC}$ and that reducing the number of parameters using the convolutional layer is important.   %

\begin{figure}[t!]
    \centering
   \includegraphics[width=1\linewidth]{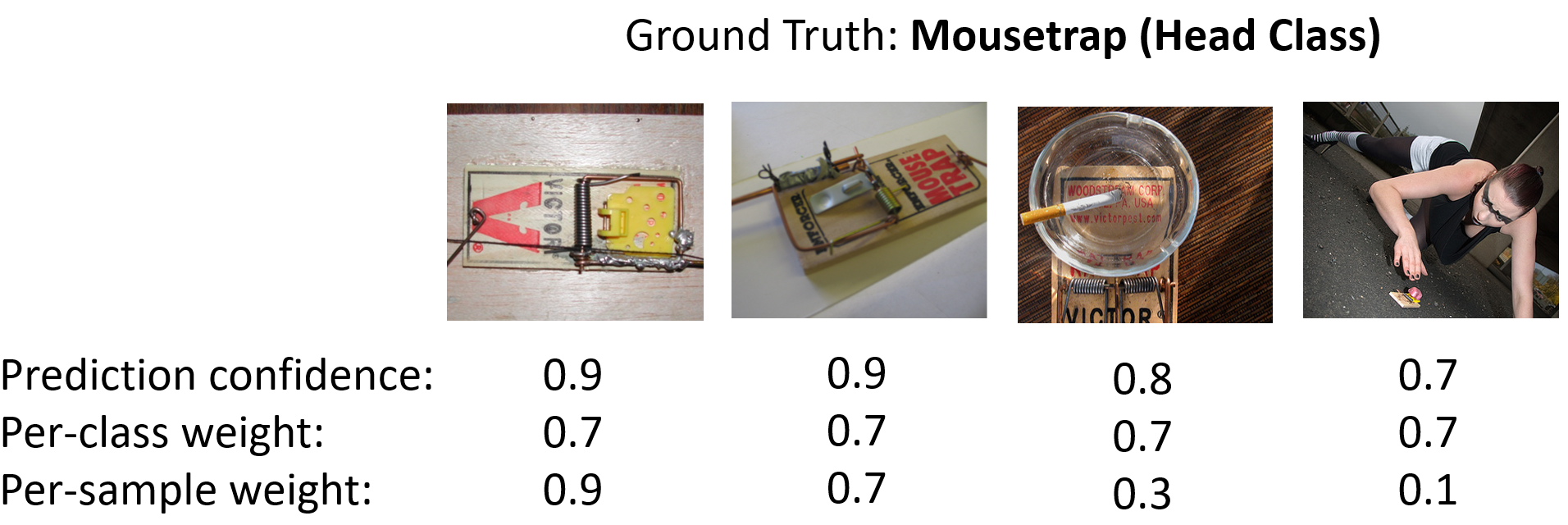}
   \caption{Using per-sample weighting, images typical for the class Mousetrap (softmax has low-entropy) are weighed more strongly than non-typical images (softmax has high-entropy). Per-class weighting reweighs all samples for that class the same (0.7), hurting recognition of typical images.}
    \label{fig:per_sample_strength}
\end{figure}

\begin{table}
\centering
    \begin{sc}
    \scalebox{0.9}{
    \setlength{\tabcolsep}{1pt} %
\begin{tabular}{lcc}
    \textbf{} & 
    \textbf{$Acc_{PC}$} &
    \textbf{$Acc_{LT}$} \\
\midrule
    Visual Expert + smDRAGON & 55.8 & 66.0 
    \\
    Semantic Expert + smDRAGON & 57.7 & 63.4
    \\
    \textbf{DRAGON (OURS)} & \textbf{60.1} & \textbf{67.7}
    \\
    \bottomrule
    \end{tabular}
    }
    \end{sc}

\caption{Ablation study, comparing sm\DRAGON{} to \DRAGON{} on CUB-LT: Fusing information between modalities improves performance  (test set, CUB-LT).}\label{tab:smDragon_test}
\end{table}

\begin{table}
\centering
      \scalebox{1}{
        \begin{tabular}{lcc}
        \textbf{Sorting} & 
        \textbf{$Acc_{PC}$} &
        \textbf{$Acc_{LT}$} \\
        \midrule
        No Sorting & 58.7 & 68.2 \\
        \textbf{Sorting By Visual Expert} & \textbf{60.0} & \textbf{70.9}\\
        Sorting By Semantic Expert & 60.0 & 70.8\\
        \bottomrule
        \end{tabular}
      }
\caption{Ablation study, quantifying the contribution of sorting the fusion-module inputs (validation set, CUB-LT).} 
\label{gater-input-train-ablation-bench}
\end{table}

To quantify the contribution of \textit{per-sample} weighting, Table \ref{tab:sample-basis} compares it against \textit{per-class} weighting on three long-tail benchmarks: ImageNet-LT, Places365-LT and CIFAR100-LT. To keep the comparison fair, this was done using vision-only ($\lambda=1$), and sweeping over the same set of hyper parameters. 
To gain more intuition on how per-sample weighting helps, Fig. \ref{fig:per_sample_strength} plots the per-sample weights of four images from a ImageNet-LT head class (mousetrap). Per-sample weighs more strongly "easy" samples (low entropy) than non-typical samples. This illustrates that per-sample weighting does not penalize ``justified" high-confidence predictions if they happen to arrive from a head class. At the same time, per-sample weighting gives more chance to tail classes, reducing the familiarity bias.

\textbf{Sharing order statistics (sorting):}
Table \ref{gater-input-train-ablation-bench} quantifies the benefit of sorting expert predictions. As discussed in \secref{appendix:more-intuition}, sorting enables sharing of information across classes by fixing the input location of each order statistic ($\max$, $2^{nd} \max$ etc.).

\section{Conclusion}
This paper discussed two key challenges for learning with long-tail unbalanced data: A ``familiarity bias", where models favor head classes, and low accuracy over tail classes due to lack of samples. We address these challenges, with \DRAGON{}, a late-fusion architecture for visual recognition that learns with per-class semantic information. It outperforms existing methods on new long-tailed versions of ImageNet, CUB, SUN, and AWA. It further sets new SoTA on a \TwoLevel{} benchmark \cite{Schnfeld2019GeneralizedZL}. Finally, a single-modality variant of \DRAGON{} improves accuracy over standard long-tail learning benchmarks, including ImageNet-LT, Places365-LT, and unbalanced CIFAR-10/100. These results show that information about the number of samples per-class can be effectively used to reduce prediction biases.

Strongly unbalanced data with a long-tail is ubiquitous in numerous domains and problems. 
The results in this paper show that a light-weight late-fusion model can be used to address many of the challenges posed by class imbalance.
\subsection*{Acknowledgments}
 \noindent DS was funded by a grant from the Israeli innovation authority, through the AVATAR consortium and by a grant from the Israel Science Foundation (ISF 737/2018). Study was also funded by an equipment grant to GC and Bar-Ilan University from the Israel Science Foundation (ISF 2332/18).

\newpage
\clearpage
{\small
\bibliographystyle{ieee_fullname}
\bibliography{dragon}
}
\clearpage

\renewcommand{\figurename}{Fig. S}
\renewcommand\tablename{Table S}
\renewcommand{\figref}[1]{Figure S\ref{#1}}
\newcommand{\tabref}[1]{Table S\ref{#1}}

\appendix

\section{Additional analysis of the familiarity effect}
\label{sec:extra-bias-effect}

Here we provide a deeper analysis showing that \DRAGON{} effectively addresses the ``familiarity bias". 

\begin{figure}[t!]
    \centering
    (a) 
    \\
    \includegraphics[width=0.6\linewidth]{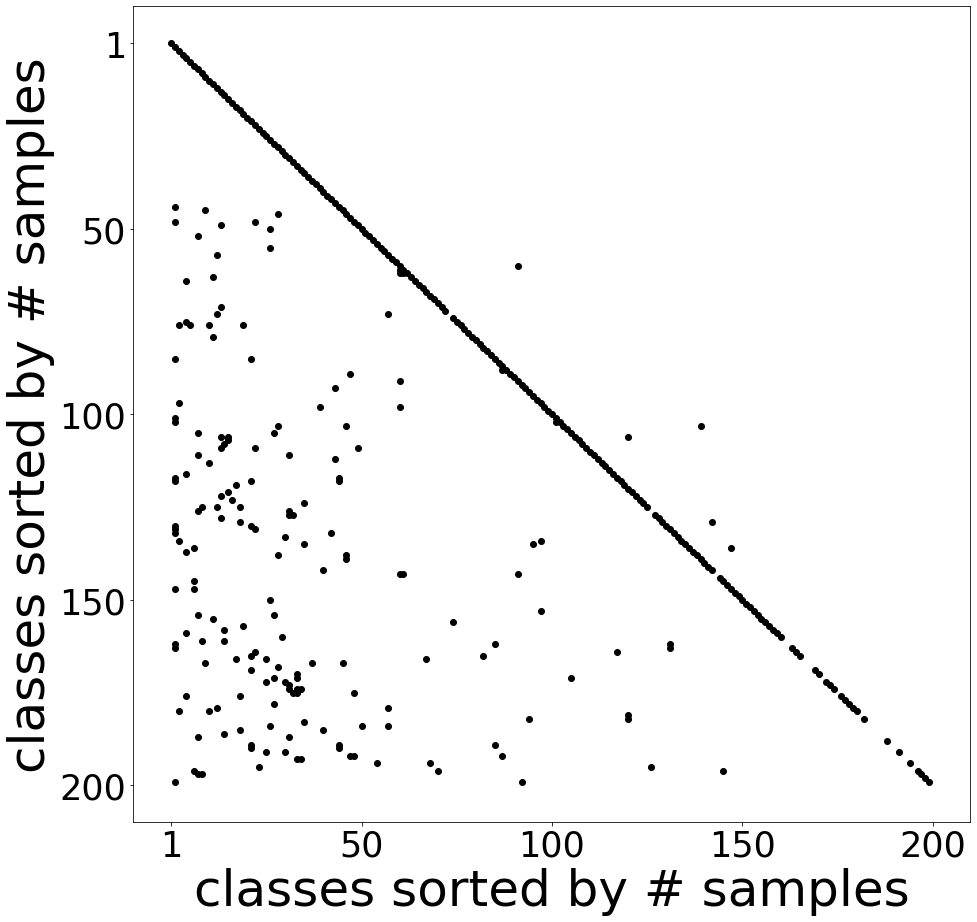} 
    \\
    (b) \hspace{80pt}
    (c)
    \\
    \includegraphics[width=0.4\linewidth]{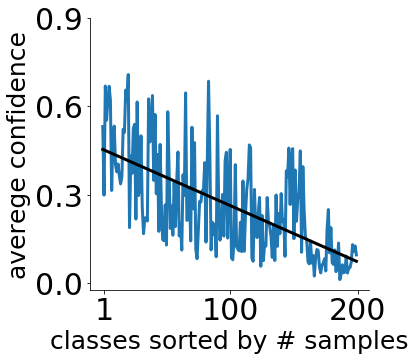}
    \includegraphics[width=0.4\linewidth]{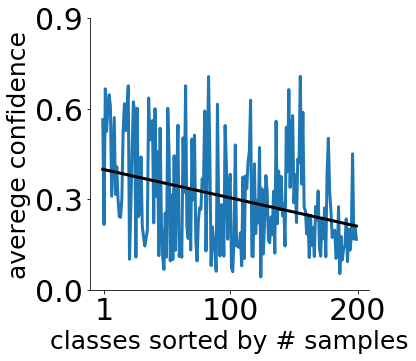}\\

    \caption{\DRAGON{} learns to offset the familiarity bias. 
    \textbf{(a)} A confusion matrix of a ResNet101 trained  
    on CUB-LT as a function of the number of samples per class. The matrix shows markers for pairs of (gt, predicted) whose count is larger than 15.
    \textbf{(b)} Average-confidence per-class of the classifier.  \textbf{(c)} Similar curve as (b) but for \DRAGON{}. 
    Black lines depict a linear regression line. \DRAGON{} per-class confidence has smaller dependence on the number of samples in the train. 
    }
    \label{fig:familiarity-bias-dragon}
\end{figure}

\textbf{The familiarity bias causes models to incorrectly favor head classes:} Figure S\ref{fig:familiarity-bias-dragon}(a) shows the confusion matrix of a standard ResNet-101 trained on CUB-LT, as computed on the validation set. Classes, of the confusion matrix, are ordered by a decreasing number of training samples, with class \#1 having many samples and class \#200 have few samples. Black dots denote count larger than 15. 

It illustrates two effects. First, the trained model correctly classifies head classes, based on the fact that the top rows have no incorrect (off-diagonal) predictions. Second, for mid and tail classes, predictions are clearly biased towards the head, since there are many more off-diagonal predictions to the left (head class predictions).

\textbf{\DRAGON{} corrects for the familiarity bias:} Figures S\ref{fig:familiarity-bias-dragon}(b) and S\ref{fig:familiarity-bias-dragon}(c) demonstrate that \DRAGON{} learns to offset the familiarity bias. 
The left panel (b) shows the familiarity effect on CUB-LT before recalibration.
The right panel (c) shows that \DRAGON{} corrects the familiarity bias and produces a more balanced average confidence across the head and tail classes.

\begin{figure}[t!]
    \centering
    (a)\hspace{80pt}
    (b)
    \\
    \includegraphics[width=0.44\linewidth]{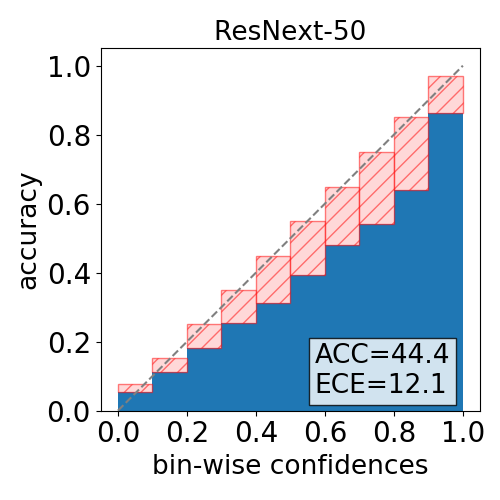}
    \includegraphics[width=0.44\linewidth]{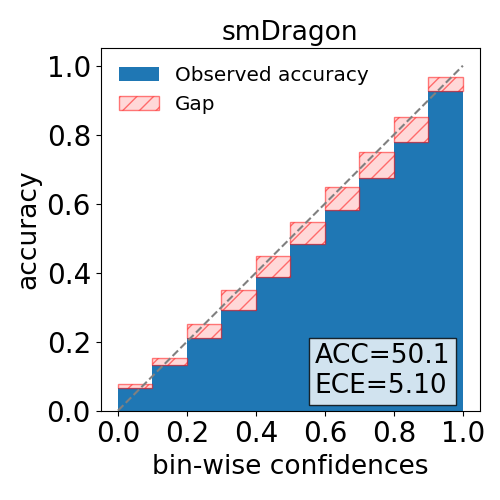}
    \\
    (c)\hspace{80pt}
    (d)
    \\
    \includegraphics[width=0.44\linewidth]{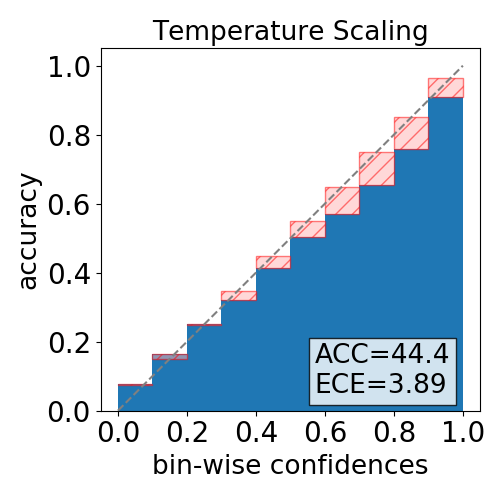}
    \includegraphics[width=0.44\linewidth]{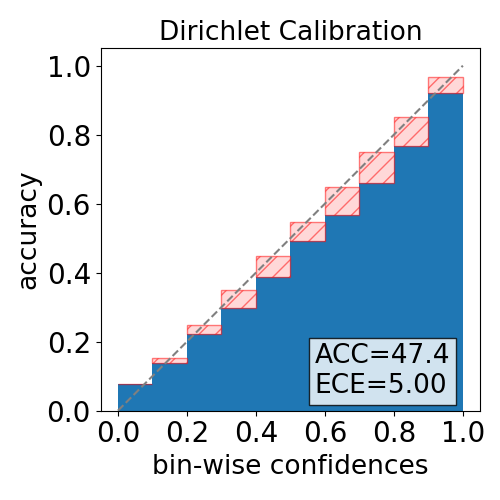}

    \caption{Reliability Diagrams on ImageNet-LT, for (a) raw ResNext-50,  (b) sm\DRAGON{},  (c) Temperature-Scaling~\cite{Guo2017OnCO} and (d)  Dirichlet-Calibration~\cite{Kull2019BeyondTS}. We report expected-calibration-error (ECE) and per-class accuracy (ACC)}
    \label{fig:reliability_plots}
\end{figure}

\textbf{\DRAGON{} re-calibrate predictions:}
In the main paper (Figure \ref{fig:unbalance}(c)) we showed that a model that is trained on unbalanced data has higher confidence for head classes and it over-estimate them. \textbf{By reversing the familiarity bias, sm\DRAGON{}, implicitly, also re-calibrate experts predictions.} Figure \ref{fig:reliability_plots} compares the reliability diagrams for sm\DRAGON{} against raw \textit{ResNeXt-152}, \textit{Temp Scaling~\cite{Guo2017OnCO}} and  \textit{Dirichlet Calibration~\cite{Kull2019BeyondTS}} (common and SoTA calibration approaches).
We report both per-class-accuracy (ACC) and expected-calibration-error (ECE) for each model.

\section{Visual experts are better at the head, 
\\
semantic experts excel at the tail}
\label{appendix:visual_vs_attribute}

Here we provide supporting evidence to our observation from Section \ref{sec:approach} of the main paper that semantic experts are better at the tail: ``\textit{Semantic descriptions of classes can be very useful for tail (low-shot) classes, because they allow models to recognize classes even with few or no training samples~\cite{DAP,xianCVPR,LAGO}}''. Additionally, we demonstrate that the visual expert is better for the many-shot regime.

We focus on the Two-Level CUB distribution, and evaluate the accuracy for the many shot classes when restricting predictions to these classes (many-among-many), and separately the accuracy for the few-shot classes when predictions are restricted to these tail classes (few-among-few). 

\begin{figure}[t!]
    \centering
    (a) Few-among-few\hspace{20pt}(b) Many-among-many \\
    \includegraphics[width=0.44\linewidth]{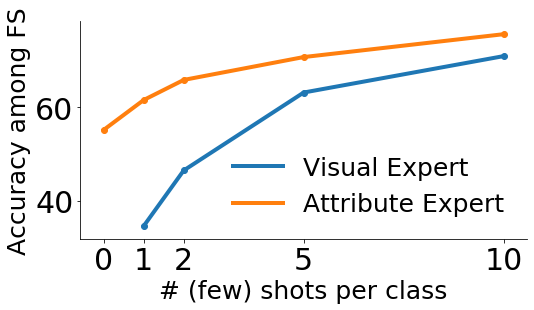}
    \includegraphics[width=0.44\linewidth]{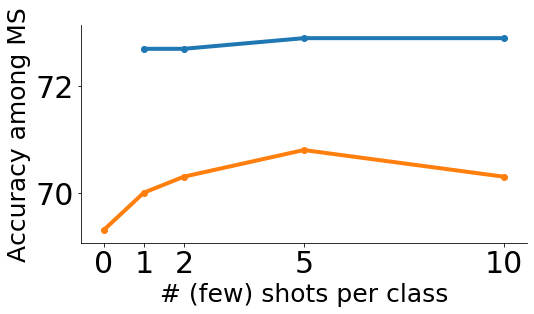}
    \caption{Accuracy as a function of number of samples at the tail, of the \textit{Visual Expert} and the \textit{Semantic Expert} used in our study.
    \textbf{(a)} Accuracy among few-shot classes; The Semantic expert outperforms the visual expert.
    \textbf{(b)} Accuracy among many-shot classes; The Visual expert outperforms, regardless of the number of samples at the tail. 
    }
    \label{fig:ve_vs_ae}
\end{figure}

\figref{fig:ve_vs_ae}(a) shows the accuracy over few-shot classes of both experts in few-among-few setting. The semantic expert outperforms the visual one, and this effect stronger with fewer samples. For example, with 1-shot learning, the semantic expert is almost 100\% better than the Visual Expert.
Additionally, when we measure the accuracy of the many-shot classes in the many-among-many setup (accuracy at the head), the visual expert is better than the semantic expert \figref{fig:ve_vs_ae}(b). 

\begin{figure}
    \centering
    (a)\hspace{100pt}(b) \\
    \includegraphics[width=0.48\linewidth]{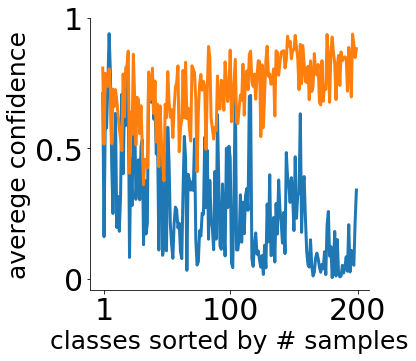}
    \includegraphics[width=0.48\linewidth]{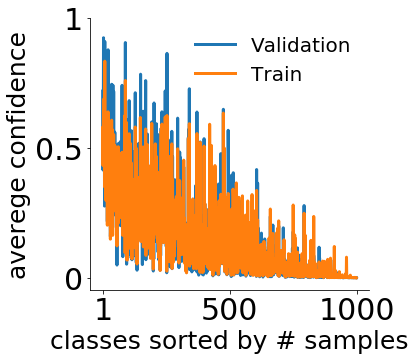}
    \caption{The familiarity bias effect: \textbf{(a)} On CUB-LT the effect is strong on validation samples (blue) but not on training samples (orange). \textbf{(b)} On ImageNet-LT the effect is prominent on both train and validation samples.  }
    \label{fig:train_vs_val_bias}
\end{figure}

\section{Training the fusion-module in small scale datasets}
\label{sec:training-fusion-module}
Our goal is to have the fusion-module learn to capture the correlations between the number of training samples and the output confidence (the familiarity bias), so it can adjust for it. 
Unfortunately, while the familiarity effect is substantial in the validation data and the test data, it may not present in the training data in small scale datasets.
The reason is: Models tend to overfit and become overconfident over rare classes in the training set. 
This effect is illustrated in \figref{fig:train_vs_val_bias}(a) (compare Train versus Validation curves) for CUB-LT. We  observed the effect in also in SUN-LT and AWA-LT. 

To address this mismatch, we hold-out 50\% of the samples of the tail classes and 20\% of the samples of the head classes of the training data and use it to simulate the response of experts to test samples. This set is used for training the fusion-module.

Note, that after training the fusion module, we re-train the experts on all the training data (including the hold-out set), in order to use all data available. (See \secref{sec:implemntation} for more details).

In large-scale datasets, like ImageNet-LT, no hold-out set is needed and \DRAGON{} is trained on the training set. There,  the familiarity bias is also present on the training data (\figref{fig:train_vs_val_bias}(b)), as the models did not overfit the tail classes.

\section{Implementation details}
\label{sec:implemntation}

\paragraph{Training:} Considering the observation from \secref{sec:training-fusion-module}, regarding CUB-LT, SUN-LT and AWA-LT, we train the architecture in three steps: 
First, we train each expert on the training data excluding the hold-out set. Second, we freeze the expert weights and train the \textit{fusion-module} on all the training set. Finally, we re-train the experts on all the training data in order to use all data available.
For the hold-out set, we randomly draw half the samples of the tail classes and 20\% of the samples of the head classes.
For inference, we use the fusion-module trained at the second step with the experts trained at the third step.  

\noindent\textbf{Platt-scaling:}
We used Platt-scaling~\cite{Platt99} to tune the combination coefficient $\lambda$ by adding constant bias $\beta$ and applying a sigmoid on top of its scores:
$\lambda = \sigma\Big[f_0 - \beta\Big]$, where 
$\beta$ is a hyperparameter selected with cross validation.

\noindent\textbf{Fusion-module:}
We trained the fusion-module using ADAM~\cite{Adam} optimizer. For large-scale datasets, like ImageNet-LT, Places-LT and iNaturalist, we used $L_2$ regularization, selected by hyperparameter optimization using grid search $\in \{10^{-5},10^{-4},10^{-3}\}]$, to avoid overfitting.
\noindent\textbf{Hyper-parameter tuning:}
We determined the number of training epochs (early-stopping), selected architecture alternatives, and tuned hyperparameters using the validation set, using $Acc_{LT}$ for\textit{ \SmoothTail{}} and \textit{Vision-only}, and $Acc_{PC}$ for \textit{\TwoLevel{}}.
\newline For \textit{DRAGON}: 
We optimized the following hyperparameters: (1) Number of filters in the convolution layer~$\in \{1,.., 4\}$. (2) Degree of polynomial in Eq.\ref{eq_polynom}~$\in \{2,3,4\}$. (3) Learning rate~$\in\{10^{-5},10^{-4},10^{-3}\}$. (4) Bias term of Platts rescaling $\beta\in [-2,2]$.
\newline\noindent For \textit{CADA-VAE~\cite{Schnfeld2019GeneralizedZL}}: We applied a grid search for the latent embedding space $\in [12,25,50,64,100,200,250]$, variational-autoencoder learning rate $\in [0.0001,0.00015,...,0.015]$ and classifier learning rate $\in [0.0001, 0.0005,...,0.1]$. We used a batch size of 64.
\newline\noindent For \textit{Focal Loss~\cite{Lin2017FocalLF}}: We applied a grid-search for gamma $\in [1,2,...,15]$ and alpha $\in [0.1,0.2,0.5,0.75,0.9,1]$.
\newline\noindent For \textit{Range Loss~\cite{Lin2017FocalLF}}: We applied a grid-search for alpha $\in [0.1,0.2,0.5,0.75,0.9,1]$ and beta $\in [0.1,0.2,0.5,0.75,0.9,1]$.
\newline\noindent For \textit{Anchor Loss~\cite{Lin2017FocalLF}}: We applied a grid-search for gamma $\in [0.1,0.5,1,...,15]$ and slack $\in [0.001,0.005,0.01,...,0.5]$.
\newline\noindent For \textit{LDAM Loss~\cite{Lin2017FocalLF}} we applied a grid-search for C $\in [0.1,0.2,...,0.9]$.

\subsection{Computing $Acc_{LT}$:} 
\label{appendix:long-tail-eval}
$Acc_{LT}$ measures the accuracy over a test distribution that resembles the training distribution. However, the test and validation samples of CUB-LT, SUN-LT and AWA-LT have a different distribution because they were originated from an approximate uniform distribution. Thus, to compute $Acc_{LT}$ we measure the accuracy for each individual class, and then take a weighted sum according to the class frequencies in the training set. Specifically, for each class, we assign a weight $P_{train}(y)$ according to the train-set distribution such that $0 < P_{train}(y) < 1$ and $\sum_y p_{train}(y) = 1$. Then we compute the accuracy per class and report the weighted average across all classes: $Acc_{LT}=\sum_{y=1}^k{p_{train}(y)acc(y)}$. This is equivalent to transforming the test set to have the same distribution as the train set.

\subsection{A clarification about the \SmoothTail{} \\ benchmark}
\label{appendix:clarification_class_alignment}
In this section, we explain how the long-tail benchmark was aligned with the two-level benchmark, as was mentioned in the paragraph that describes the long-tailed datasets (Section \ref{sec:smooth-tail} of the main paper).

To align the long-tail benchmark with the two-level benchmark, we first ordered the classes according to their number of samples in the \textit{two-level} distribution. Then we calculated the number of samples for each class according to the required long-tail distribution, and accordingly drew samples to construct the training set.

\begin{table}[t!]
    \begin{center}
      \scalebox{1}{ %
    \setlength{\tabcolsep}{2pt} %
\hskip-.5cm
    \begin{tabular}{lcc}
    \textbf{Sorting} & 
    \textbf{$Acc_{PC}$} &
    \textbf{$Acc_{LT}$}  \\
\midrule
    No-Sorting & 58.5 & 57.0
    \\
    \textbf{Sorting-By-Visual-Expert} & \textbf{60.1} & \textbf{67.7} 
    \\
    Sorting-By-Semantic-Expert & 59.8 & 67.7
    \\
    \bottomrule
    \end{tabular}
    }
    \end{center}
     \caption{Ablation study, quantifying the contribution of sorting the fusion-module inputs (test set, CUB-LT).}
  \label{sort-ablation-bench-test}
\end{table}

\begin{table}[t!]
    \begin{center}
     \scalebox{1}{
    \setlength{\tabcolsep}{4pt} %
    \begin{tabular}{lcc}
    \textbf{Architecture} & 
    \textbf{\(Acc_{PC}\)} &
    \textbf{\(Acc_{LT}\)} \\
    \midrule
    F.C. & 56.4 & 66.3     \\
    F.C. $\&$ $1/n_y$ re-scale & 56.4 & 56.2    \\
    F.C. $\&$ non-parametric re-scale & 58.2 & 64.3    \\
    Conv. $\&$ non-parametric re-scale & 58.4 & 67.1    \\
    Conv. $\&$ single parametric re-scale & 59.3 & 64.3    \\
    \textbf{\DRAGON{} (ours)} & \textbf{60.1} & \textbf{67.7}
    \\
    \bottomrule
    \end{tabular}}
    \end{center}
  \caption{Ablation study, comparing different fusion and re-scaling approaches.
  The results show the contribution of the convolutional backbone and the re-scaling method for the two experts (test set, CUB-LT).}
  \label{architecture-ablation-bench-test}
\end{table}

 \begin{table}[t!]
    \begin{center}
      \scalebox{1}{ %
    \setlength{\tabcolsep}{2pt} %
\hskip-.5cm
    \begin{tabular}{lc}
    \textbf{Training Process} & 
    \textbf{$Acc_{PC}$}   \\
    \midrule
    All-Train & 56.6     \\
    End-To-End & 46.4     \\
    \textbf{Three-Stage-Training} & \textbf{60.1}    \\
    \bottomrule
    \end{tabular}
    }
    \end{center}
     \caption{Ablation study, quantifying the contribution the effect of three-stage training as proposed in Section \ref{sec:implemntation}. (test-set, CUB)}
  \label{training-ablation-bench-test}
\end{table}

\begin{table*}
\begin{subtable}{\textwidth}
\centering
   \begin{tabular}{l|c|c|c}
   Model & 
    \textbf{$Acc_{ms}$} & \textbf{$Acc_{fs}$} &
    \textbf{$Acc_{H}$} \\
  \midrule
    Most Common Class* &
    0.7, 0.7, 0.7, 0.7 & %
    0, 0, 0, 0 & %
    0, 0, 0, 0  %
    \\
    \hline
    LDAM~\cite{cao2019learning}* &
    71.5, 71.9, 71.6, 71.5 & %
    1.2, 5.9, 24.1, 41.2 & %
    2.4, 10.9, 36.0, 52.2   %
    \\
    \hline
    REVISE~\cite{REVISE} &
    - & %
    - & %
    36.3, 41.1, 44.6, 50.9  %
    \\
    CA-VAE~\cite{Schnfeld2019GeneralizedZL}&
    58.2, 57.6, 60.0, 62.2 & %
    44.8, 51.6, 59.4, 62.3 & %
    50.6, 54.4, 59.6, 62.2   %
    \\
    DA-VAE~\cite{Schnfeld2019GeneralizedZL} &
    50.6, 56.0, 56.8, 56.8 & %
    47.9, 53.2, 61.0, 65.4 & %
    49.2, 54.6, 58.8, 60.8   %
    \\
    CADA-VAE~\cite{Schnfeld2019GeneralizedZL} &
    59.6, 60.9, 62.3, 63.1 & %
    51.4, 57.5, 63.6, 68.8 & %
    55.2, 59.2, 63.0, 64.9  %
    \\
    \hline
    CE Loss* \textbf{(VE)} &
    72.7, 72.9, 72.7, 72.0 & %
    0.6, 3.7, 19.1, 38.6 & %
    1.2, 6.9, 30.2, 50.2   %
    \\
    LAGO~\cite{LAGO}* \textbf{(SE)} &
    69.2, 69.0, 69.0, 68.1 & %
    13.8, 21.9, 38.1, 51.5 & %
    23.0, 33.2, 49.0, 58.6   %
    \\
    \textbf{\DRAGON{} (ours)} &
    58.0, 62.9, 63.3, 66.1 & %
    52.8, 55.9, 63.8, 69.6 & %
    \textbf{55.3, 59.2, 63.5, 67.8}   %
    \\
\bottomrule
\end{tabular}
\caption{Two-Level CUB}
   \label{tab:sub_first}
\end{subtable}

\bigskip
\begin{subtable}{\textwidth}
\centering
   \begin{tabular}{l|c|c|c}
   Model & 
    \textbf{$Acc_{ms}$} & \textbf{$Acc_{fs}$} &
    \textbf{$Acc_{H}$} \\
  \midrule
    Most Common Class* &
    0.2, 0.2, 0.2, 0.2 & %
    0, 0, 0, 0 & %
    0, 0, 0, 0  %
    \\
    \hline
    LDAM~\cite{cao2019learning}* &
    43.7, 44.0, 44.2, 44.3 & %
    2.2, 6.6, 19.0, 31.8 & %
    4.3, 11.5, 26.6, 37.0
    \\
    \hline
    REVISE~\cite{REVISE} &
    - & %
    - & %
    27.4, 33.4, 37.4, 40.8  %
    \\
    CA-VAE~\cite{Schnfeld2019GeneralizedZL} &
    35.8, 37.5, 37.5, 39.0 & %
    40.0, 46.5, 53.8, 55.7 & %
    37.8, 41.4, 44.2, 45.8 %
    \\
    DA-VAE~\cite{Schnfeld2019GeneralizedZL} &
    34.8, 37.3, 38.6, 38.2 & %
    41.4, 45.1, 50.2, 54.8 & %
    37.8, 40.8, 43.6, 45.1 %
    \\
    CADA-VAE~\cite{Schnfeld2019GeneralizedZL} &
    37.6, 38.2, 39.4, 41.9 & %
    44.1, 49.0, 55.3, 55.1 & %
    40.6, 43.0, 46.0, 47.6  %
    \\
    \hline
    CE Loss* \textbf{(VE)} &
    46.3, 46.3, 46.2, 45.6 & %
    0.9, 4.9, 17.2, 33.0 & %
    1.8, 8.9, 25.1, 38.3   %
    \\
    LAGO~\cite{LAGO}* \textbf{(SE)} &
    30.6, 30.4, 30.7, 31.0 & %
    14.4, 18.7, 21.9, 25.2 & %
    19.6, 23.2, 25.6, 27.8   %
    \\
    \textbf{\DRAGON{} (ours)} &
    37.2, 39.2, 40.5, 41.6 & %
    45.5, 49.6, 55.1, 57.2 & %
    \textbf{41.0, 43.8, 46.7, 48.2}   %
    \\
\bottomrule
\end{tabular}

\caption{Two-Level SUN}
   \label{tab:sub_second}
\end{subtable}

\bigskip
\begin{subtable}{\textwidth}
\centering
   \begin{tabular}{l|c|c|c}
   Model & 
    \textbf{$Acc_{ms}$} & \textbf{$Acc_{fs}$} &
    \textbf{$Acc_{H}$} \\
  \midrule
    Most Common Class* &
    2.5, 2.5, 2.5, 2.5 & %
    0, 0, 0, 0 & %
    0, 0, 0, 0  %
    \\
    \hline
    LDAM~\cite{cao2019learning}* &
    90.7, 90.7, 90.5, 90.5 & %
    6.6, 14.4, 26.6, 41.6 & %
    12.4, 24.8, 41.1, 57.0   %
    \\
    \hline
    REVISE~\cite{REVISE} &
    - & %
    - & %
    56.1, 60.3, 64.1, 67.8  %
    \\
    CA-VAE~\cite{Schnfeld2019GeneralizedZL}&
    73.4, 77.7, 81.0, 81.0 & %
    56.8, 66.0, 72.8, 77.1 & %
    64.0, 71.3, 76.6, 79.0 %
    \\
    DA-VAE~\cite{Schnfeld2019GeneralizedZL} &
    74.0, 74.6, 73.5, 73.9 & %
    63.0, 71.4, 77.7, 79.8 & %
    68.0, 73.0, 75.6, 76.8 %
    \\
    CADA-VAE~\cite{Schnfeld2019GeneralizedZL} &
    76.6, 79.4, 81.9, 82.6 & %
    63.8, 68.7, 74.8, 78.0 & %
    \textbf{69.6, 73.7, 78.2,} 80.2  %
    \\
    \hline
    CE Loss* \textbf{(VE)} &
    90.7, 90.9, 89.7, 87.9 & %
    5.9, 11.2, 32.6, 57.9 & %
    \\
    LAGO~\cite{LAGO}* \textbf{(SE)} &
    82.6, 81.9, 81.7, 81.5 & %
    11.5, 20.6, 46.2, 59.4 & %
    20.2, 33.0, 59.0, 68.7   %
    \\
    \textbf{\DRAGON{} (ours)} &
    74.5, 76.7, 79.2, 81.7 & %
    61.1, 62.9, 74.3, 82.1 & %
    67.1, 69.1, 76.7, \textbf{81.9}   %
    \\
\bottomrule
\end{tabular}
\caption{Two-Level AWA}
   \label{tab:sub_third}
\end{subtable}
\caption{Comparing  \DRAGON{} with SoTA GFSL models and baselines with increasing number of few-shot training samples on the CUB, SUN and AWA datasets. We report per-class $Acc_{ms}$, $Acc_{fs}$ and  $Acc_{H}$. Each cell represents 1-shot,2-shot,5-shot and 10-shot accuracies}
   \label{tab:three_tables}
\end{table*}

\subsection{Training CADA on \SmoothTail{} benchmark}
\label{appendix:training-cada}
In this section, we explain how we trained CADA-VAE~\cite{Schnfeld2019GeneralizedZL} for the long-tail benchmark.

To evaluate CADA-VAE~\cite{Schnfeld2019GeneralizedZL} on long-tail benchmarks we used the code published by the authors and followed the training protocol exactly as they used for the two-level distribution. Since the protocol relies on a hard distinction between head classes and tail classes, we had to choose where to partition the smooth long-tail distribution to head and tail. Our solution is simple. It is based on the fact that we aligned the order of classes in the long-tail distribution to be the same order as in the two-level split (\secref{appendix:clarification_class_alignment}). The alignment allowed us to use the same partition to head and tail as used for the two-level benchmark.

\section{Additional metrics}

\subsection{$Acc_{fs}$ and $Acc_{ms}$ on Two-Level benchmarks}
\label{appendix:ms-fs-two-level}
\tabref{tab:three_tables} provides the results of $Acc_{fs}$ and $Acc_{ms}$ (described in section \ref{sec:two-level}) for the Two-Level benchmark. We show results for 1,2,5 and 10-shots.
At the main paper we reported the results for the $Acc_H$ metrics, which is derived from $Acc_{fs}$ and $Acc_{ms}$ reported here.

\subsection{Ablation results on the test set}
\label{appendix:ablation-test-set}

In the main paper (\ref{sec:ablation}) we described results for ablation study on the validation set. Here we report results for the same model variant on the test set. 

Tables S\ref{sort-ablation-bench-test} and S\ref{architecture-ablation-bench-test} show the results of the ablation study on the test set. It shows the same behavior as the ablation study on the validation set that was reported in the main paper.
Table S\ref{training-ablation-bench-test} compares three different training protocols: (1) \textit{All-Train}: Training the \DRAGON{} fusion-module naively without a hold-out set. 
(2) \textit{End-To-End}: Training all the architecture  (both experts and fusion-module) end to end in an early fusion manner. 
(3) \textit{Three-Stage-Training}: Training our models as explained in section \ref{sec:implemntation}.

\clearpage

\end{document}